\documentclass[a4paper,fleqn]{cas-sc} %

\usepackage[authoryear]{natbib}

\def\tsc#1{\csdef{#1}{\textsc{\lowercase{#1}}\xspace}}
\tsc{WGM}
\tsc{QE}
\tsc{EP}
\tsc{PMS}
\tsc{BEC}
\tsc{DE}

\usepackage{amsmath}
\usepackage{graphicx}
\usepackage{subcaption}

\usepackage{placeins}
\usepackage{lineno}
\usepackage{makecell}
\usepackage{longtable}
\makeatletter
\newcommand\subsubsubsection{\@startsection{paragraph}{4}{\z@}{-2.5ex\@plus -1ex \@minus -.25ex}{1.25ex \@plus .25ex}{\normalfont\normalsize\bfseries}}
\newcommand\subsubsubsubsection{\@startsection{subparagraph}{5}{\z@}{-2.5ex\@plus -1ex \@minus -.25ex}{1.25ex \@plus .25ex}{\normalfont\normalsize\bfseries}}
\makeatother

\usepackage{boxedminipage}
\usepackage{textpos}

\begin{document}
\let\WriteBookmarks\relax
\def\floatpagepagefraction{1}
\def\textpagefraction{.001}
\shorttitle{}

\title [mode = title]{Assessing bikeability with street view imagery and computer vision}

\author[1]{Koichi Ito}%
\author[1,2]{Filip Biljecki}%
\cormark[1]
\ead[url]{filip@nus.edu.sg}

\address[1]{Department of Architecture, National University of Singapore, Singapore}
\address[2]{Department of Real Estate, National University of Singapore, Singapore}

\cortext[cor1]{Corresponding author}

\begin{abstract}
Studies evaluating bikeability usually compute spatial indicators shaping cycling conditions and conflate them in a quantitative index.
Much research involves site visits or conventional geospatial approaches, and few studies have leveraged street view imagery (SVI) for conducting virtual audits.
These have assessed a limited range of aspects, and not all have been automated using computer vision (CV).
Furthermore, studies have not yet zeroed in on gauging the usability of these technologies thoroughly.
We investigate, with experiments at a fine spatial scale and across multiple geographies (Singapore and Tokyo), whether we can use SVI and CV to assess bikeability comprehensively.
Extending related work, we develop an exhaustive index of bikeability composed of 34 indicators.
The results suggest that SVI and CV are adequate to evaluate bikeability in cities comprehensively.
As they outperformed non-SVI counterparts by a wide margin, SVI indicators are also found to be superior in assessing urban bikeability and potentially can be used independently, replacing traditional techniques.
However, the paper exposes some limitations, suggesting that the best way forward is combining both SVI and non-SVI approaches.
The new bikeability index presents a contribution in transportation and urban analytics, and it is scalable to assess cycling appeal widely.
\end{abstract}


\maketitle

\begin{textblock*}{\textwidth}(0cm,-11.45cm)
\begin{center}
\begin{footnotesize}
\begin{boxedminipage}{1\textwidth}
This is the Accepted Manuscript version of an article published by Elsevier in the journal \emph{Transportation Research Part C: Emerging Technologies} in 2021, which is available at: \url{https://doi.org/10.1016/j.trc.2021.103371}. Cite as:
Ito K, Biljecki F (2021): Assessing bikeability with street view imagery and computer vision. \textit{Transportation Research Part C}, 132: 103371.
\end{boxedminipage}
\end{footnotesize}
\end{center}
\end{textblock*}

\begin{textblock*}{\textwidth}(-0.5cm,14.12cm)
{\tiny{\copyright{ }2021, Elsevier. Licensed under the Creative Commons Attribution-NonCommercial-NoDerivatives 4.0 International (\url{http://creativecommons.org/licenses/by-nc-nd/4.0/})}}
\end{textblock*}

\section{Introduction}

\begin{figure}[tbp]
     \centering
     \begin{subfigure}[b]{.49\textwidth}
         \centering
         \includegraphics[width=\textwidth]{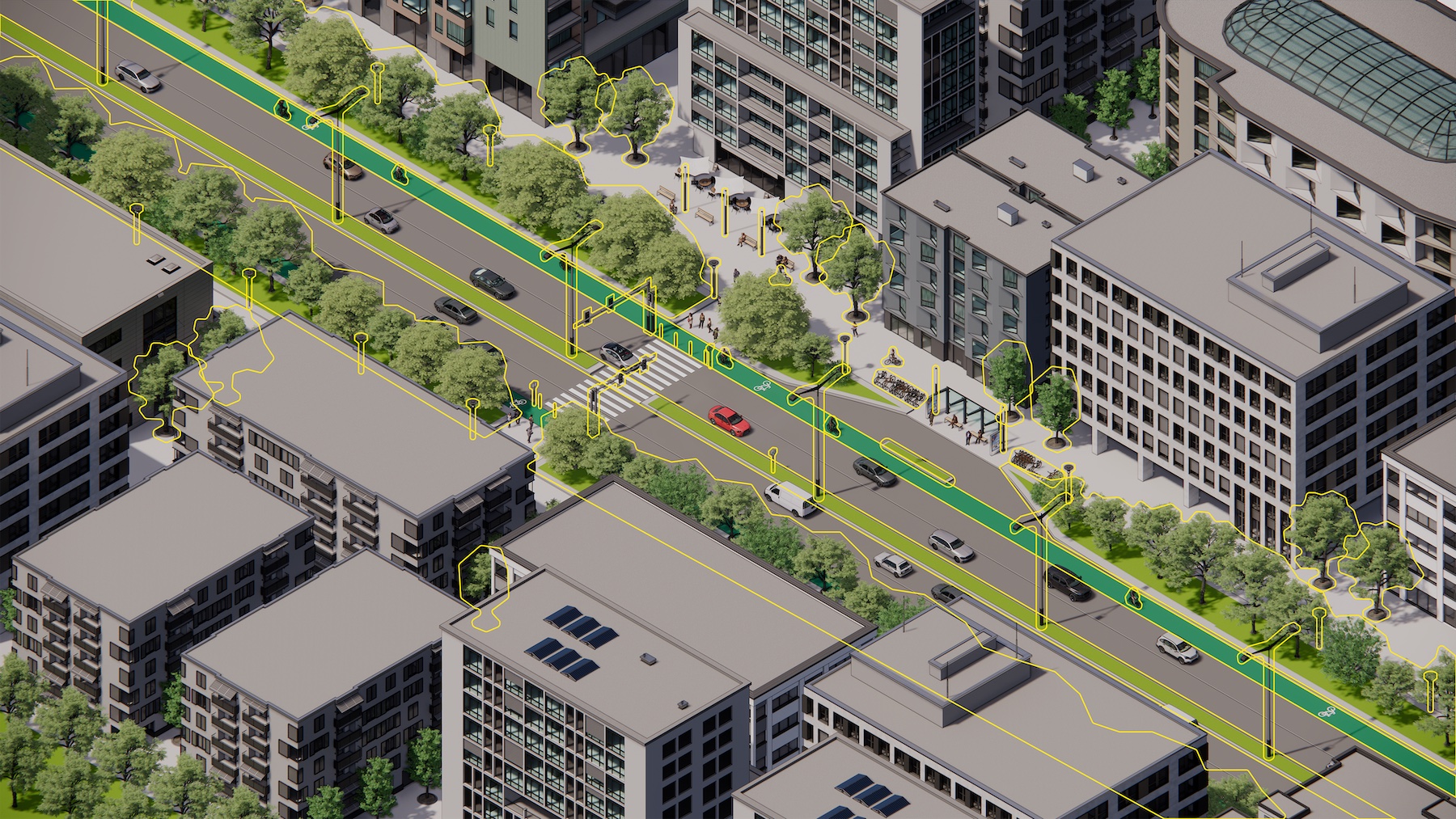}
     \end{subfigure}
     \begin{subfigure}[b]{.49\textwidth}
         \centering
         \includegraphics[width=\textwidth]{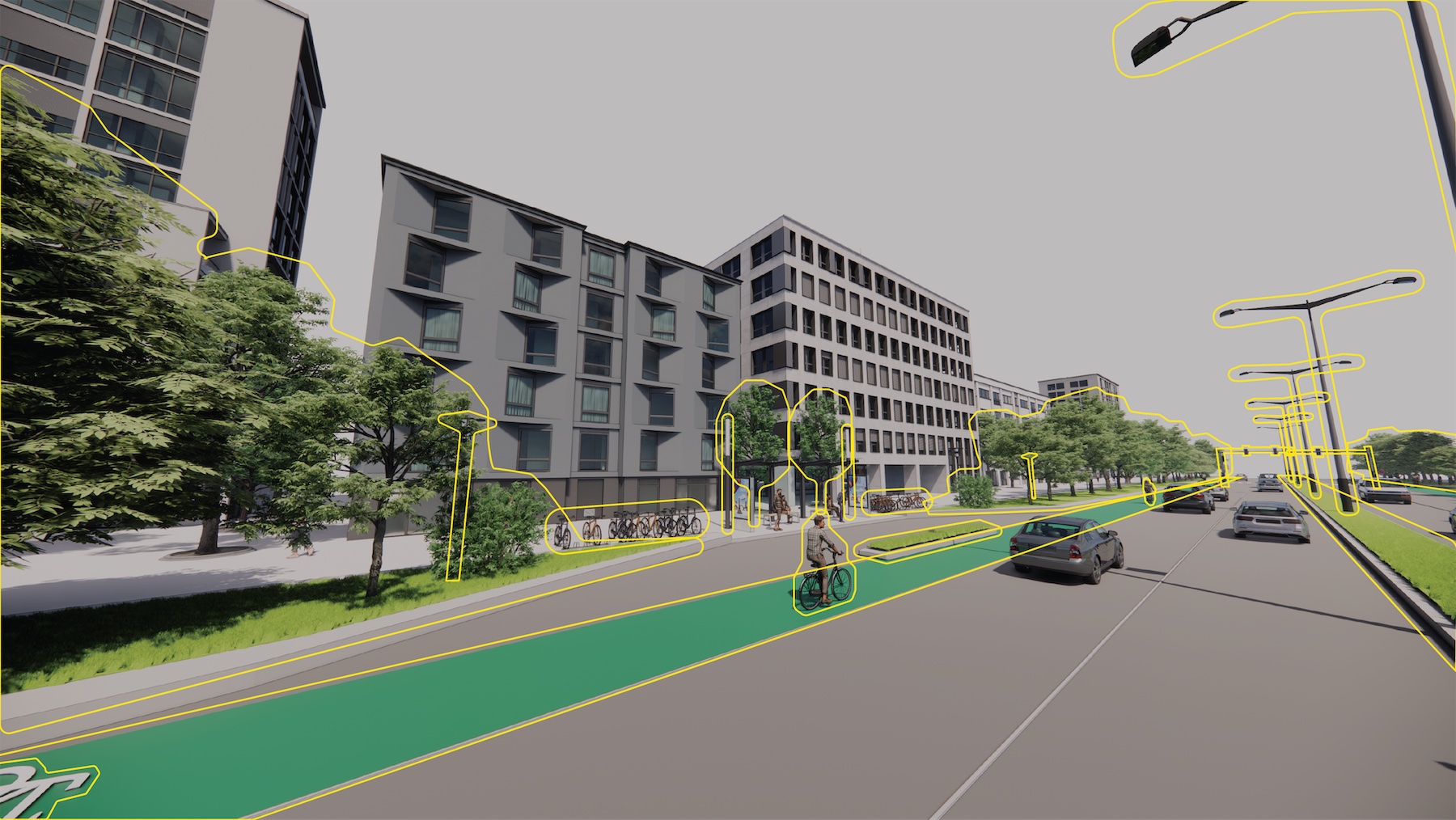}
     \end{subfigure}
        \caption{Illustration of an urban setting together with one of the corresponding street-level views, highlighting several aspects that may indicate bikeability. The method presented in this paper takes advantage of a substantial number of visual features that may be extracted automatically from street view images and engage them to generate a composite index that suggests cycling appeal at a fine spatial scale and across multiple cities.}
        \label{methodology:fig:street_view_example}
\end{figure}

Bicycles play an important role in making cities environmentally sustainable, healthy, and economically vibrant \citep{neves_assessing_2019,Cao_2019,rojas-rueda_health_2011,volker_economic_2021,wang_modeling_2016,Horacek2018578,wang_commute_2019, Yeh2019,Chen2020,Abdellaoui_Alaoui_2021}. 
To evaluate the extent to which cycling is facilitated, the notion of bikeability was created by complementing the concept of walkability.
Subsequently, index systems, as instruments to quantify it, have been developed by many studies \citep{porter_bikeability_2020,cain_development_2018,manton_using_2016, gullon_assessing_2015,koh_influence_2013, horacek_sneakers_2012,wahlgren_active_2010,hoedl_bikeability_2010, clifton_development_2007,titze_developing_2012,arellana_developing_2020,winters_mapping_2013,guler_location_2021, Lin2018381,Gholamialam201973,Resch20201,Kamel2020,Grigore2019607,Osama2020767,Chevalier202059,Kang2019,Schmid-Querg20211,Galanis2018461, lowry_low-stress_2016,faghih_imani_cycle_2019,boongaling_developing_2021}.
This topic became even more relevant with the advent of bicycle-sharing systems \citep{Du_2019,Luo_2020}.
Data collection methodologies to calculate bikeability indexes are for a large part derived from field observations, entailing time-consuming manual work and limiting the spatial extent that can be surveyed.
As technologies have advanced, these assessments have been supplemented by new data sources, such as crowdsourcing and virtual observations \citep{kalvelage_assessing_2018,gullon_assessing_2015,Abadi_2018}.
Nevertheless, previous studies face various issues, such as slow data collection process, the balance between subjectivity and objectivity of data, lack of street-level information, and standardization of spatial granularity. 
The recent availability of street view imagery (SVI) has yielded opportunities for new approaches to urban studies~\citep{Biljecki.2021}, enabling a wealth of images from the pedestrian and cyclist perspective that may be used to assess walkability and bikeability and that is available remotely (\autoref{methodology:fig:street_view_example}).
In parallel, developments in computer vision (CV) have catalyzed the means to process the profusion of photos automatically and efficiently.
They have already been utilized for assessing walkability \citep{nagata_objective_2020}.
SVI and CV are not entirely new to bikeability either.
For example, studies by \citet{tran_cyclists_2020} and \citet{gu_using_2018} have used CV and SVI to assess particular aspects of bikeability.
However, no comprehensive bikeability assessment study has been conducted yet, and there has been no critical evaluation of the usability of such technologies in comparison to conventional methods that have been dominating the field so far.

Considering the developments in computer vision and proliferation of street view imagery (e.g.\ increased coverage of commercial services such as Google Street View and introduction of crowdsourced alternatives), we believe that such research is needed and timely.
This study tests a hypothesis that CV and SVI coupled together have a strong potential to overcome the issues faced by conventional methods in gauging how friendly urban streets are to cyclists.
Further, as the availability of SVI is now considered to be virtually omnipresent, a notable research gap is comparative studies that involve more than one city.
Thus, in this study we aim to answer the following research questions: can we use CV techniques and SVI data to comprehensively assess bikeability within \textit{and} among cities?
If yes, can SVI and CV alone be used to assess bikeability, replacing traditional techniques entirely?
To answer these questions, we develop a bikeability index with 34 indicators under five categories and implement it in Singapore and Tokyo.
The contributions of our study are the novel investigation of CV techniques' and SVI indicators' usefulness in comprehensive bikeability assessment, construction of a new comprehensive bikeability index regarding an unprecedented amount of aspects, and provision of a new data collection method that can extract subjective and objective indicators from street-level information at a larger spatial scale and across multiple geographies, thereby overcoming the issues found by the previous studies.
We believe that the method and the index may be scaled around the world, including additional cities in future work.
It is important to note that our method includes also a survey with a large number of human participants to investigate whether CV techniques may estimate human perception of cycling appeal automatically.

In Section~\ref{sc:background}, we conduct a comprehensive literature review to affirm the research gap.
The comprehensive and structured overview of the state of the art is another contribution of this paper.
Section~\ref{sc:methodology} explains the data sources and methodologies used in this study.
Section~\ref{sc:overview} describes the results of this study together with a discussion.
Finally, we conclude our study as well as discuss further directions for future research in Section \ref{sc:conclusion}.

\section{Related work}\label{sc:background}
\subsection{Bikeability studies}
Many studies have explored various aspects of the built environment that can influence people's cycling behavior \citep{bauman_correlates_2012, Nielsen201836, Pritchard2019, daraei_data-driven_2021,kraus_provisional_2021,nazemi_studying_2021,mcneil_bikeability_2011,ma_peoples_2017, cicchino_not_2020,Nogal20201,Sottile2019543, Berger2018524,Porter2018,Aldred2020,Long202055,Doubleday2021,Bruchert20201,Martin2021,Attard2021501}.  
Studies on the association of the built environment and cycling conditions became well-established, and many researchers developed indexes to assess specific aspects of the built environment that can affect cycling behavior and comprehensively quantify bikeability, i.e.\ the extent to which an environment is friendly for bicycling. 

In the early days of related work between 2010 and 2015, methodologies to collect data are mostly conducted through field surveys, thus tending to be time- and resource-intensive. 
Studies by \citet{hoedl_bikeability_2010}, \citet{horacek_sneakers_2012}, and \citet{koh_influence_2013} involve field observation by experts, and such a methodological limitations precludes bikeability assessment at large scales. 
Moreover, only a few studies \citep{koh_influence_2013} include both objective and subjective indicators, and most of the studies \citep{wahlgren_active_2010,hoedl_bikeability_2010,horacek_sneakers_2012} only focused on either one of the indicators.
The spatial granularity of sample points is loosely defined and not standardized in some studies as well \citep{wahlgren_active_2010,horacek_sneakers_2012}.

Recent bikeability studies have become more standardized and scalable.
Although some studies still use field observation \citep{manton_using_2016,cain_development_2018}, more studies apply emerging technologies and data sources such as remote sensing images, manual virtual auditing using SVI, and crowdsourcing, to collect data \citep{krenn_development_2015, gullon_assessing_2015,winters_bike_2016,kalvelage_assessing_2018}, enabling large-scale and comparative assessment of bikeability.
However, as data collection methods become more scalable, subjective indicators are excluded by many studies \citep{krenn_development_2015, manton_using_2016,winters_bike_2016,cain_development_2018,kalvelage_assessing_2018}.
Moreover, remotely sensed imagery cannot capture street-level information, and manual virtual auditing and crowdsourcing data collection require a large amount of time and resources. 
For example, a recent study by \citet{arellana_developing_2020} utilizes virtual auditing in SVI, but this data collection process was reported to be six months long for a city-scale study area, suffering from the same issue of a time-intensive method as the aforementioned studies.
To overcome the issue of the balance between street-level information and scalability, recent studies couple SVI with CV techniques to automate the indicator extraction \citep{gu_using_2018,tran_cyclists_2020}, but the number of indicators extracted by such methods remains still limited.
Moreover, subjective indicators for bikeability assessment have not been extracted from SVI by using CV yet.
Therefore, there is a need for further studies to examine the possibility of extracting more indicators --- both objective and subjective --- from SVI by using CV.

Structuring the rundown on related work, \autoref{background:tab:indicators} summarizes indicators used in the reviewed studies.
Most of the indicators could be categorized into connectivity, environment, infrastructure, vehicle--cyclist interaction (V--C interaction), and perception.
In developing bikeability indexes, the previous studies have faced the issues of the time-intensive data collection process, the balance between subjectivity and objectivity, extraction of street-level information, and standardization of spatial granularity. 
\autoref{background:tab:data_weight_automate_street_info} summarizes the issues mentioned above.
Literature reviews on bikeability by \citet{kellstedt_scoping_2021} and \citet{castanon_bikeability_2021} indicate that the past development of bikeability assessment has been driven by innovative uses of advanced new technologies, thereby suggesting that newer technologies may overcome these issues mentioned above.
While there have been studies that have utilized SVI and CV techniques to assess bikeability, and thus indicating the reliability of SVI as a data source that can be used in a scalable manner, they suffer from shortcomings that we seek to mitigate in our work.
Primarily, previous studies used these technologies to assess very limited aspects of bikeability (i.e.\ only up to a few indicators), did not collect and assess subjective indicators, and have been evaluated on limited areas.
Such a gap necessitates further studies to examine how much SVI and CV techniques are usable to assess bikeability comprehensively.

\subsection{Street view imagery in urban studies}%
The growth of the spatial coverage of street view imagery and the development of computer vision techniques have catalyzed the recent proliferation of studies that utilize them, both in transportation and urban planning and beyond \citep{Wang_2019,Song:2020,Wu.2021pdi,Ye.2020,Fan.2021,Chen.2021r5p}.
This section focuses on studies that use SVI to extract information of data used in previous bikeability studies under four categories that have been delineated by this study (i.e.\ environment, infrastructure, vehicle-cyclist interaction, and perception), and to examine subjects pertaining to cycling. 

One of the more explored aspects in related research is urban greenery. 
Quantification of greenery by using image segmentation, CV techniques that can classify categories of objects at a pixel-level, has enabled many various urban studies ranging from simple assessment of the distribution of vegetation and interdisciplinary examinations of the relationships between greenery and various aspects of cities, such as physical activities of residents and real estate \citep{ye_measuring_2019,lu_using_2019,ye_daily_2019}. 
Other studies
quantify greenery, sky view factor (i.e.\ openness), and buildings (i.e.\ enclosure) with semantic segmentation to measure characteristics of cities in a scalable manner \citep{li_quantifying_2017, gong_mapping_2018, gong_classifying_2019, dacci_using_2019, toikka_green_2020, wang_life_2020, ma_measuring_2021, zhou_using_2021}. 
These features that are extracted from SVI, which have been leveraged for a variety of applications, can be also used to evaluate bikeability. 

The high scalability of CV techniques and SVI has also multiplied opportunities for infrastructure assessment at a city scale. 
\citet{hall_traffic_2018} propose a methodology to detect and classify traffic signals, and \citet{chacra_municipal_2018} develop a CV-based model to detect infrastructure anomalies from SVI. 
Assessment of urban accessibility is conducted by \citet{najafizadeh_feasibility_2018}, developing models to detect accessibility problems from SVI, such as missing curb ramps and street surface issues.
Further, \citet{ding_towards_2021} use object detection and classification CV models to map bike lane networks from SVI.
Mapillary, a crowdsourced SVI service, developed a dataset called Vistas, which contains 25,000 images collected from around the world, and annotated according to 66 categories \citep{neuhold_mapillary_2017}.
Cityscapes is another frequently used street-level dataset for segmentation \citep{cordts_cityscapes_2016, gong_classifying_2019, nagata_objective_2020}.
Although Mapillary Vistas has many infrastructure-related categories (e.g.\ bike lane and bike parking) that other segmentation training datasets do not regard, it has not been widely used in urban studies.
Thus, using this dataset might expand the prospects of SVI for bikeability assessment.  

These techniques have been utilized in transport studies as well. 
\citet{goel_estimating_2018} find that the number of cyclists manually counted in GSV images is strongly correlated with the cycling mode share reported by cities in the UK (r = 0.92). 
\citet{zhang_social_2019} and \citet{chen_estimating_2020} also examine the relationship between visual features and urban traffic volume and reveal that SVI can be used to explain more than 65 percent of the spatiotemporal mobility pattern. 
These findings hint that SVI may be used to estimate traffic volume, which is an important aspect of bikeability \citep{Labetski.2020}.

\begin{table}[!ht]
\centering
\scriptsize
\caption{An overview of indicators used by previous studies grouped into five categories.}
\label{background:tab:indicators}
\begin{tabular}{lllllll}\toprule
Publication &Connectivity &Environment &Infrastructure &V-C interaction &Perception \\\midrule
\citet{clifton_development_2007}&Cul-de-sac &Land use &Sidewalk & &Attractiveness \\
&Continuity &Slope &Pavement &Traffic speed &Safety \\
& &Greenery &Path obstruction &Street parking &Cleanliness \\
& &Enclosure &Sidewalk buffer &Traffic control & \\
& &Building design &Road condition & & \\
& &Setback &Curb cuts & & \\
& & &Crossing & & \\
& & &Street light & & \\
& & &Street amenity & & \\
& & &Directional sign & & \\
& & &Power line & & \\
& & &Bike lane & & \\
& & &Transit facilities & & \\
\citet{hoedl_bikeability_2010}&N/A &Greenery &Bikelane & &N/A \\
& &Land use &Sidewalk &Traffic speed & \\
& &Billboards & &Traffic volume & \\
& &Open space & & & \\
\citet{wahlgren_active_2010}&Directness &Air quality &N/A &Traffic speed &Attractiveness \\
&Intersection &Noise & &Traffic volume &Crowdedness \\
& &Greenery & &Cylist speed &Safety \\
& &Slope & &Traffic separation &Beauty \\
\citet{horacek_sneakers_2012} &N/A &Slope &Pavement &Traffic speed &Beauty \\
& & &Street light &Traffic volume & \\
& & &Potholes &Traffic control & \\
& & &Path size & & \\
& & &Sidewalk buffer & & \\
& & &Curb cut & & \\
& & &Bike lane & & \\
\citet{koh_influence_2013}&Detour &Slope &Directional sign &N/A &Safety \\
&Intersection &POIs &Pavement & &Crowdedness \\
& &Greenery &Shelter & & \\
& &Water & & & \\
\citet{gullon_assessing_2015} &Different routes &Greenery &Pavement & &Cleanliness \\
& &Land use mix &Street amenity &Traffic control &Beauty \\
& & &Street light & &Attractiveness \\
& & &Crossing & & \\
\citet{krenn_development_2015} &N/A &Greenery &Bike lane &Traffic separation &N/A \\
& &Water & & & \\
& &Slope & & & \\
\citet{manton_using_2016}&Intersection &N/A &Road width &Traffic volume &N/A \\
& & & &Street parking & \\
& & & &Traffic separation & \\
\citet{winters_bike_2016} &Intersection &Slope &Bike lanes &Mode share &N/A \\
&Distance to POIs & & & & \\
\citet{hartanto_developing_2017} &Intersection &Water &Road type &Traffic speed &N/A \\
&Directness &Greenery &Pavement &Traffic volume & \\
& &Buildings &Street light & & \\
& &Slope &Bike parking & & \\
\citet{cain_development_2018}&Intersection &Land use &Transit facilities &Traffic control &N/A \\
&Informal path &No. of pedestrian &Roll-over curb & & \\
&Cul-de-sac &Water &Street amenity & & \\
& &Landscape &Bike parking & & \\
& &Hardscape &Curb cuts & & \\
& &Graffiti &Bike lane & & \\
& &Setback &Crossing & & \\
& &Shade &Road width & & \\
& &Building design &Sidewalk buffer & & \\
& & &Bike lane & & \\
& & &Sidewalk & & \\
\citet{gu_using_2018} &Street density &Shade &Bike lane &Traffic separation &N/A \\
& &POIs &Crossing & & \\
\citet{arellana_developing_2020} &N/A &Slope &Bike lane &Traffic control &N/A \\
& &Greenery &Sidewalk &Mode share & \\
& &Building design &Street obstruction &Traffic speed & \\
& &Crime presence &Bike lane width & & \\
& & &Street light & & \\
& & &Police presence & & \\
& & &Security camera & & \\
\citet{tran_cyclists_2020}&Directness &POIs &Bike lane &N/A &N/A \\
& &Greenery & & & \\
& &Enclosure & & & \\
& &Crowdedness & & & \\
\bottomrule
\end{tabular}
\end{table}

\clearpage %

\begin{table}[!htp]
\centering
\caption{Characteristics of methods used by previous studies.}\label{background:tab:data_weight_automate_street_info}
\begin{tabular}{lllrll}\toprule
\textbf{Publication} &\textbf{Data} &\textbf{Weighting system} &\textbf{\makecell[l]{No.\ of indicators\\ by SVI+CV}} &\textbf{\makecell[l]{Street-level \\information}} \\\midrule
\citet{clifton_development_2007}&Objective \& &Individual assessment &0 &Yes \\
&Subjective & & & \\
\citet{hoedl_bikeability_2010}&Objective &Individual assessment &0 &Yes \\
\citet{wahlgren_active_2010}&Subjective &Individual assessment &0 &Yes \\
\citet{horacek_sneakers_2012}&Objective &Unequal arbitrary weight &0 &Yes \\
\citet{koh_influence_2013}&Objective \& &Survey-based weight &0 &Yes \\
&Subjective & & & \\
\citet{gullon_assessing_2015}&Objective \& &Unequal arbitrary weight &0 &Yes \\
&Subjective & & & \\
\citet{krenn_development_2015} &Objective &Equal weight &0 &No \\
\citet{manton_using_2016} &Objective &Regression modeling &0 &Yes \\
\citet{winters_bike_2016}&Objective &Equal weight &0 &No \\
\citet{hartanto_developing_2017} &Objective &Equal weight &0 &Yes \\
\citet{cain_development_2018} &Objective &Unequal arbitrary weight &0 &Yes \\
\citet{gu_using_2018}&Objective &Entropy weight &3 &No \\
\citet{arellana_developing_2020}&Objective &Survey-based weight &0 &Yes \\
\citet{tran_cyclists_2020}&Objective &Equal weight &3 &Yes \\
\bottomrule
\end{tabular}
\end{table}

Extraction of information from images has also enabled the prediction of urban perception based on SVI.
\citet{naik_streetscore_2014} conduct surveys on safety, in which 7,872 unique participants from 91 countries ranked 4,109 images using 208,738 pairwise comparisons. 
\citet{dubey_deep_2016} develop a dataset called Place Pulse 2.0, which consist of 110,988 images from 56 cities and 1,170,000 pairwise comparisons answered by 81,630 online survey participants on six perception scores, and predicted these scores with convolutional neural network models, such as VGGNet, and this study's dataset and methodology have been utilized and replicated by other studies \citep{kang_human_2021,qiu_subjectively_2021}.
\citet{yao_discovering_2021} conduct a similar study to predict perception scores by surveying 20 volunteers on 1,000-2,000 images and predicting perception scores using a random forest model and features extracted by FCN-8s as inputs.
\citet{verma_predicting_2020} also recruit only 79 participants to rate 200 images, extracting high- and low-level features from SVI by using CV techniques (e.g.\ image segmentation, object detection, classification, and edge detection), and designing models to predict fourteen perception scores, such as ``pleasant'', ``boring'', and ``safe''. 
Moreover, the validity of conducting perception surveys has been examined and found to be as reliable as surveys based on the real environment by \citet{feng_validity_2021}.

\citet{lu_associations_2019} and \citet{wang_relationship_2020} examine the association between urban greenery (extracted from SVI) and cycling behaviors, and \citet{hollander_using_2020} study the correlation between transportation planning and the perceived safety of the built environment. 
A study by \citet{tran_cyclists_2020} is the only study that relies on SVI to create a bikeability index, but this study uses SVI only to assess very limited aspects of bikeability, such as greenery and enclosure.
Therefore, there has not been any study that developed a comprehensive bikeability index mirroring the wide array of aspects jointly covered by related work and one that has utilized SVI as a major data source to calculate it, a gap that is bridged by our study.

\subsection{Summary}%
The literature review elucidates the proliferation of urban studies that utilize SVI and CV techniques and, at the same time, it exposes the absence of studies that take advantage of them for assessing cycling appeal and developing a bikeability index. 
Such omission is possibly caused by obstacles of using CV techniques and the complexities of developing a comprehensive bikeability index system.
However, previous studies demonstrate the possibility of replacing traditional data collection for bikeability indicators, as they collect information on similar aspects that may indicate cycling appeal.

This paper aims to bridge this gap by designing a thorough bikeability index that is largely derived using SVI and CV, and it also uses the opportunity to critically investigate their value and independence when doing so.
Moreover, as SVI data has virtually global coverage, the study puts scalability under the spotlight; thus, the combination of scalability of SVI and efficient extraction of indicators through CV can reduce the time and resource cost of bikeability assessment compared to conventional data sources and methods.
Finally, potential barriers of this approach for urban planners are also examined to understand its application and prospects for adoption.

\section{Methodology}\label{sc:methodology}
This study uses six data sources (i.e.\ SVI, surveys, OpenStreetMap (OSM), Land Use (LU), Digital Elevation Model (DEM), and Air Quality Index (AQI)) to evaluate 34 indicators under five categories (i.e.\ connectivity, environment, infrastructure, perception, and vehicle-cyclist interaction).
The comprehensive selection of these indicators is a contribution on its own, as its scope is unprecedented, and may serve as a resource to develop indexes in other domains.
Testing the scalability of the method, Singapore and Tokyo, as two geographies with disparate characteristics, are selected as study areas.
Thus, data retrieval and indicator extractions are conducted to develop the composite index by weighting each category and indicator equally.
\autoref{methodology:fig:street_view_example} highlights some examples of phenomena in SVI that are characterized in this method.

\subsection{Selection of indicators}  %
Conducting a review of studies that developed bikeability indexes, this study devised an exhaustive list of indicators used by them \citep{hartanto_developing_2017,winters_bike_2016, gullon_assessing_2015,wahlgren_active_2010,horacek_sneakers_2012,hoedl_bikeability_2010,clifton_development_2007,manton_using_2016,cain_development_2018,koh_influence_2013}. 
This inventory was used to create an own index to expand related work and minimize the bias when selecting indicators.
Duplicates and those indicators that cannot be obtained and/or are unsuitable for this study have been excluded, such as noise and the presence of informal paths and crimes.
After filtering, out of 65 unique indicators found in the previous studies in total, 34 indicators (i.e.\ about 52\%) were kept, forming the most extensive instance to date.
Future studies can use our method while partially modifying this list to incorporate more or fewer indicators as long as they can be extracted from the data sources we used.

These 34 unique indicators are categorized into five categories, namely, connectivity, environment, infrastructure, perception, and vehicle-cyclist interaction (see \autoref{methodology:tab:index_table}).
Regarding data sources, 21 indicators are to be extracted from SVI, 10 from OSM, one from each of LU, DEM, and AQI.

Although this study does not examine if the selected indicators can predict bike usage, most of the reviewed previous studies also selected them based on literature reviews, except for a few \citep{winters_bike_2016,arellana_developing_2020} that use regression analysis.
Moreover, higher bikeability does not necessarily lead to higher counts of bike usage.
For example, \citet{arellana_developing_2020} examine and find that other factors, such as population density and socio-economic characteristics, also play a role \citep{munira_estimating_2021}.
Therefore, investigating whether the bikeability index can predict bike usage is out of the scope of this research.

\autoref{methodology:tab:index_table} lists indicators with their data sources, extraction methods, and scaling methods.
For scaling methods, min-max scaling and negative min-max scaling were used (see \autoref{methodology:equation:min_max} and \autoref{methodology:equation:neg_min_max}).
The sampling method is further explained in the following sections.
\begin{equation}
x_{\text {Min-Max Scaled}}=\frac{x-\min (x)}{\max (x)-\min (x)}
\label{methodology:equation:min_max}
\end{equation}
\begin{equation}
x_{\text {Negative Min-Max Scaled}}=1-\frac{x-\min (x)}{\max (x)-\min (x)}
\label{methodology:equation:neg_min_max}
\end{equation}

\subsection{OSM and SVI data retrieval} %
This study relies on OSM to retrieve information on indicators and also for the retrieval of locations on streets that are used to fetch street view images.
The completeness of OSM in the study area is deemed adequate \citep{BarringtonLeigh:2017ie,Biljecki.2020}.
To retrieve the OSM data, OSMnx \citep{boeing_osmnx_2017} was used, obtaining about 250,000 points and 440,000 points for Singapore and Tokyo, respectively.
After that, 7,142 points were randomly selected.
The number of points to be collected was set at 7,142 because this is the maximum number of SVI images that can be retrieved within the free credit provided by GSV API every month.
Future studies can also utilize the initial trial credit as well, but if there are financial constraints, the rise of volunteered SVI, such as KartaView and Mapillary, may ameliorate this issue in the future.
For this study, the API of GSV was used to collect images because of its extensive coverage and high quality in both metropolises.

\begin{table}[!htp]\centering
\caption{Indicators forming the comprehensive bikeability index introduced by this study.}\label{methodology:tab:index_table}
\scriptsize
\begin{tabular}{lllll}\toprule
\textbf{Indicators} &\textbf{Data} &\textbf{Extraction } &\textbf{Scale} \\\midrule
\textbf{Connectivity} & & & \\
No.\ of intersection with lights &OSM &Aggregation (500m) &0-1 (Negative Min-Max) \\
No.\ of intersection without lights &OSM &Aggregation (500m) &0-1 (Negative Min-Max) \\
No.\ of cul-de-sac &OSM &Aggregation (500m) &0-1 (Negative Min-Max) \\\midrule
\textbf{Environment} & & & \\
Slope &DEM &Calculation &0-1 (Negative Min-Max) \\
No.\ of POI &OSM &Aggregation (500m) &0-1 (Min-Max) \\
Shannon land use mix index &Land use &Aggregation (500m) &0-1 (Min-Max) \\
Air quality index &AQI &Spatial interpolation &0-1 (Negative Min-Max) \\
Scenery: greenery &SVI &Segmentation &0-1 (Min-Max) \\
Scenery: buildings &SVI &Segmentation &0-1 (Min-Max) \\
Scenery: water &SVI &Segmentation &0-1 (Min-Max) \\\midrule
\textbf{Infrastructure} & & & \\
Type of road &OSM &Aggregation (100m) &service, track = 0.1 \\
& & &primary, primary\_link = 0.2 \\
& & &secondary, secondary\_link = 0.4 \\
& & &tertiary, tertiary\_link = 0.5 \\
& & &unclassified = 0.6 \\
& & &pedestrian path = 0.8 \\
& & &cycleway = 1 \\
& & &Others = 0 \\
Presence of potholes &SVI &Segmentation &1 if not present \\
& & &0 if present \\
Presence of street light &SVI &Segmentation &1 if present \\
& & &0 if not present \\
Presence of bike lanes &SVI &Segmentation &1 if present \\
& & &0 if not present \\
No. of transit facilities &OSM &Aggregation (500m) &0-1 (Min-Max Scale) \\
Type of pavement &OSM &Aggregation (100m) &unhewn\_cobblestone, cobblestone = 0.2 \\
& & &sett, metal, wood = 0.4 \\
& & &paved = 0.5 \\
& & &concrete:lanes, plates, paving\_stones = 0.6 \\
& & &asphalt, concrete = 1 \\
& & &Others = 0 \\
Presence of street amenities &SVI &Segmentation &1 if present \\
& & &0 if not present \\
Presence of utility pole &SVI &Segmentation &1 if not present \\
& & &0 if present \\
Presence of bike parking &SVI &Segmentation &1 if present \\
& & &0 if not present \\
Road width &OSM &Aggregation (100m) &0-1 (Divide by 10) \\
& & &1 if width is larger than 10m \\
Presence of sidewalk &SVI &Segmentation &1 if present \\
& & &0 if not present \\
Presense of crosswalk &SVI &Segmentation &1 if present \\
& & &0 if not present \\
Presence of curb cuts &SVI &Segmentation &1 if present \\
& & &0 if not present \\\midrule
\textbf{Perception} & & & \\
Attractiveness for cycling &SVI &Surveys, Modeling &0-1 (Min-Max) \\
Spaciousness &SVI &Surveys, Modeling &0-1 (Min-Max) \\
Cleanliness &SVI &Surveys, Modeling &0-1 (Min-Max) \\
Building design attractiveness &SVI &Surveys, Modeling &0-1 (Min-Max) \\
Safety as a cyclist &SVI &Surveys, Modeling &0-1 (Min-Max) \\
Beauty &SVI &Surveys, Modeling &0-1 (Min-Max) \\
Attractiveness for living &SVI &Surveys, Modeling &0-1 (Min-Max) \\\midrule
\textbf{Vehicle-Cyclist Interaction} & & & \\
No.\ of vehicles &SVI &Detection, Aggregation (500m) &0-1 (Negative Min-Max) \\
Presence of on-street parking &OSM &Aggregation (100m) &1 if not present \\
& & &0 if present \\
Presence of traffic lights / stop signs &SVI &Segmentation &1 if present \\
& & &0 if not present \\
No. of speed control devices &OSM &Aggregation (100m) &1 if present \\
& & &0 if not present \\
\bottomrule
\end{tabular}
\end{table}

\clearpage

\subsection{Extraction of indicators}\label{sc:SVI-method} %
The feature extraction process for indicators differs among different categories and data sources.
Feature extractions for land use, topography, and AQI are relatively straightforward.
For land use, we use an entropy formula derived from the Shannon index developed by \citet{frank_linking_2005} within 500m buffers from sample points (see \autoref{methodology:equation:shannon_index}).

\begin{equation}
LUM=-1\left(\sum_{i=1}^{n} p_{i} * \ln (p_{i})\right) / \ln (n)
\label{methodology:equation:shannon_index}
\end{equation}

In this formula, $LUM$ is the land-use mix score, $pi$ is the proportion of the neighborhood covered by the land use $i$ against the total area for all the land-use categories, and $n$ is the number of land-use categories. A land-use mix score of 1 indicates the highest mix possible while a score of 0 indicates the area contains a single land use. This results in values between 0, the lowest mixed-use level, and 1, the highest mixed-use level.

For topography, the slope is calculated from DEM data created by \citet{yamazaki_high-accuracy_2017} and sampled at each sample point, and the value was scaled with min-max scaling.

AQI data is collected from the Air Quality Index\footnote{\url{https://aqicn.org/here/}} by automating the retrieval of stations and AQI data in both cities with a Python library called Selenium.
The annual average AQI of stations in the cities is calculated by taking the daily maximum value of various pollutants (e.g.\ PM2.5) in the unit of µg/m3, and then spatial interpolation was conducted to estimate the average AQI at the sample points in this study by using inverse distance weighted interpolation (see \autoref{methodology:equation:idw}). 

\begin{equation}
z_{p}=\frac{\sum_{i=1}^{n}\left(\frac{z_{i}}{d_{i}^{p}}\right)}{\sum_{i=1}^{n}\left(\frac{1}{d_{i}^{p}}\right)}
\label{methodology:equation:idw}
\end{equation}

In this formula, $z_{p}$ denotes the interpolated values of the target points, $n$ represents the number of points used to interpolate from, $z_{i}$ shows the value being interpolated from, and $d_{i}^{p}$ is the distance between the target point and the point being used to interpolate from. 
The calculated AQI is then scaled into values between 0 and 1 by using min-max scaling.

Feature extractions from OSM are conducted by taking buffers from sample points.
For the density of intersection with/without lights, cul-de-sac, shops along the route, and transit facilities, we created 500m buffers from sample points and aggregated the number of these indicators because we need to consider the surrounding contexts as well. 
Other indicators collected from OSM, such as the type of roads and the type of pavement, are collected by creating 100m buffers from sample points and converting categorical values to numerical if necessary.

Finally, this study conducts feature extractions from SVI in two ways.
For objective indicators, two CV techniques --- segmentation and object detection  --- were used to extract features. 
For segmentation, the In-Place Activated BatchNorm model trained on Mapillary Vistas with WideResNet38 and DeepLab3 developed by \citet{bulo_-place_2018} is selected because of its high accuracy of the mean intersection over union of 53.42\%.
Indicators such as the scenery along bike lanes (i.e.\ built-up area, greenery, sky view factor, water) are quantified by calculating the ratios of the pixels categorized as them over the total number of pixels in the image, which are then scaled into between 0 and 1 with min-max scaling. 
Because it is not meaningful to calculate pixels of other objects such as street lights, bike lanes, street amenities, and bike parking, they are quantified in a binary manner (i.e.\ score 1 if they are in the image and 0 if they are not in the image).
Object detection is used for one indicator, that is, the number of vehicles, as a previous study suggests that SVI can be used to estimate the traffic in the area \citep{zhang_social_2019}.
For this task, we opt for the GluonCV's model zoo developed by \citet{guo_gluoncv_2020} and select a YOLOv3 model pre-trained on Pascal VOC dataset with Darknet53 as the base model, which can detect the number of bicycles, pedestrians, and vehicles with an average precision of 58.2\% at the intersection over a union of 0.5. 
We chose this model because of its reported relatively high speed and accuracy compared to other models, such as Faster Region-based Convolutional Neural Networks and Single Shot Detection \citep{srivastava_comparative_2021,li_agricultural_2020}.
Moreover, the core of our paper is separate from the models, thus if necessary, new models can be used in any part of this methodology, as our approach is model-agnostic. 
Because a street view image captures only the traffic volume of the road segment at that moment, it is not entirely reliable to estimate the traffic based on just one photo at a point.
Therefore, a buffer of 500m was generated for each point, aggregating the number of vehicles found in the plentiful imagery in the buffer.
After this processing, the mode share was also scaled into values between 0 and 1 by min-max scaling.
A previous study \citep{chen_estimating_2020} uses a similar method to estimate the number of pedestrians and obtains Cronbach's alpha above 0.8, indicating SVI's high reliability as a data source.
However, it should be noted that the number of images within each buffer varies because the sample points were randomly selected, which might produce some bias. 
Also, duplication of vehicles detected in SVI can cause bias because some vehicles could have driven together with the vehicles collecting the street view images.
The aforementioned study \citep{chen_estimating_2020} did not consider this issue and still obtained high reliability; thus, we leave improvement in methodology to detect the same vehicles with models such as Siamese-convolutional neural network for future studies. 

Subjective indicators are under the perception category, including attractiveness for cycling, spaciousness, cleanliness, building design attractiveness, and perception of safety as a cyclist, and these indicators were predicted by using features extracted from SVI.
We follow the methodology used by \citet{verma_predicting_2020}, where high- and low-level features of images are used to predict perceptions of images (see 
\autoref{methodology:tab:features}).
Low-level features can be extracted through edge detection, blob detection, and Hue-Saturation-Lightness (HSL) extraction. 
Edges are quantified by detecting edges and calculating the ratio of pixels that are categorized as edges over the total number of pixels.
Blob detection was used to calculate the number of blobs in SVI, and HSL extraction is conducted to calculate the average and standard deviation of each hue, saturation, and lightness. 
It should be noted that lightness might have some bias because of variances in the time when the images were collected, although GSV has a standardized data collection procedure and post-processing of images \citep{google_street_2018}.
As for high-level features, image classification (IC), object detection (OD), and semantic segmentation (SS) are used to extract them.
For OD and SS, this study uses the same models mentioned above, and a ResNet50 model trained on Places365 data with an accuracy of 85.07\% is used for IC \citep{zhou_places_2018}. 
These extracted features are dependent variables to predict the perception indicators.

\begin{table}[!htp]\centering
\caption{Features for predicting perception.}\label{methodology:tab:features}
\begin{tabular}{lll}\toprule
\textbf{Visual features} &\textbf{Definitions} \\\midrule
tree\_ss &\% of pixels classified as trees. \\
sky\_ss &\% of pixels classified as the sky. \\
street\_ss &\% of pixels classified as street and sidewalks. \\
built\_ss &\% of pixels classified as buildings. \\
others\_ss &\% of pixels classified as other remaining outdoor classes. \\
nature &\% of pixels classified as natural elements such as sky, tree, and water. \\
shannon &Shannon entropy values calculated on SS task. \\
slum\_ic &Probability of being classified as Slum/Alley/Junkyard in IC task. \\
market\_ic &Probability of being classified as Bazaar/Flea market/Market in IC task. \\
built\_other\_ic &Probability of being classified as Downtown/Embassy/Plaza in IC task. \\
green\_other\_ic &Probability of being classified as Forest path/Forest road in IC task. \\
bicycle\_od &No. of bicycles detected in OD task. \\
bus\_od &No. of buses detected in OD task. \\
car\_od &No. of cars detected in OD task. \\
motorcycle\_od &No. of motorcycles/scooters detected in OD task. \\
person\_od &No. of persons detected in OD task. \\
traffic\_light\_od &No. of traffic lights detected in OD task. \\
truck\_od &No. of trucks/auto rickshaws detected in OD task. \\
canny\_edge\_llf &\% of pixels detected as edges. \\
no\_of\_blobs\_llf &No. of blobs. \\
hue\_mean\_llf &The mean value of the hue dimension in HSL color space. \\
hue\_std\_llf &The standard deviation of the hue dimension in HSL color space. \\
lightness\_mean\_llf &The mean value of the lightness dimension in HSL color space. \\
lightness\_std\_llf &The standard deviation of the lightness dimension in HSL color space. \\
saturation\_mean\_llf &The mean value of the saturation dimension in HSL color space. \\
saturation\_std\_llf &The standard deviation of the saturation dimension in HSL color space. \\
\bottomrule
\end{tabular}
\end{table}

To collect the training data set and build models to predict indicators suggesting perception, a survey is conducted on Amazon Mechanical Turk, for which this study has received an exemption from the Institutional Review Board of the National University of Singapore, and in which the participants were compensated financially.
For each city, 400 SVIs, i.e.\ 800 in total, were randomly selected for the survey, which was designed to have at least eight different participants rate the images on the five indicators on a scale of 0 to 10. 
The large number of participants to rate each image ensures reliability.
The occasional and inevitable outliers among the responses were detected with the median absolute deviation method and removed when the output is above three (see 
\autoref{methodology:equation:mad}).

\begin{equation}
\operatorname{MAD}=\operatorname{median}\left(\left|X_{i}-\tilde{X}\right|\right)
\label{methodology:equation:mad}
\end{equation}

$X_{i}$ denotes each observation, and \emph{\~{X}} represents the median of all the observations.

The collected dataset is split into training and validation data sets in an 80:20 ratio to build predictive models with LightGBM \citep{ke_lightgbm_2017}, which is gaining momentum in urban studies \citep{chen_classification_2021} for its high accuracy and low computational cost in training \citep{zhang_lightgbm_2019,deng_pdrlgb_2018}. 
In this study, we tuned the following hyperparameters with 10-fold cross-validation: num\_leaves, max\_depth, min\_child\_samples, min\_child\_weight, subsample, colsample\_bytree.
After conducting the prediction for the rest of the points, the predicted perception indicators are scaled into values between 0 and 1 with min-max scaling. 

Moreover, this study explores the relationships between all the features and perception scores by grouping observations into below- and above-average values of features and conducting Welch's t-test \citep{delacre_why_2017}. 
We aim to reveal the underlying visual effects of each feature on human perceptions.

\subsection{Development of a new composite index to assess bikeability}
The composite index of this study, one of its advancements and key contributions, is developed based on conflating bikeability indexes developed by previous studies. 
There are three types of weighting systems developed by them, which warrant a brief overview.
One of them is the independent assessment of indicators.
This method only evaluates each indicator but does not give weights to them, and it is used by \citet{wahlgren_active_2010} and \citet{hoedl_bikeability_2010}.
Another type is the arbitrary weight.
This method gives arbitrary weights to categories and indicators, which is adopted by \citet{cain_development_2018,horacek_sneakers_2012}.
The last type is equal weight.
This method gives equal consideration to all the categories and indicators, which actually --- strictly speaking --- belongs to the arbitrary weighting system, but this study differentiated them for clarity.
This method is used by \citet{hartanto_developing_2017}, \citet{winters_bike_2016}, and \citet{tran_cyclists_2020}, and it is a common system among the reviewed studies.
This study adopts the equal weight system for its simplicity and scaled all the indicators into values between 0 and 1 to prevent any indicators from excessively influencing the composite index at the end (see \autoref{methodology:equation:bikeability_index}).
\begin{equation}
Index=\sum_{i=1}^{n} \left(x_{i} * (100/(N_{c} * N_{c i}))\right)
\label{methodology:equation:bikeability_index}
\end{equation}
In this equation, \emph{Index} represents the bikeability index, $x_{i}$ denotes the value of each indicator $i$, $N_{c}$ show the number of categories, and $N_{c i}$ stand for the number of indicators in the respective category.

\begin{figure}[!t]
  \centering
  \scalebox{0.2}{\includegraphics[width=5\linewidth]{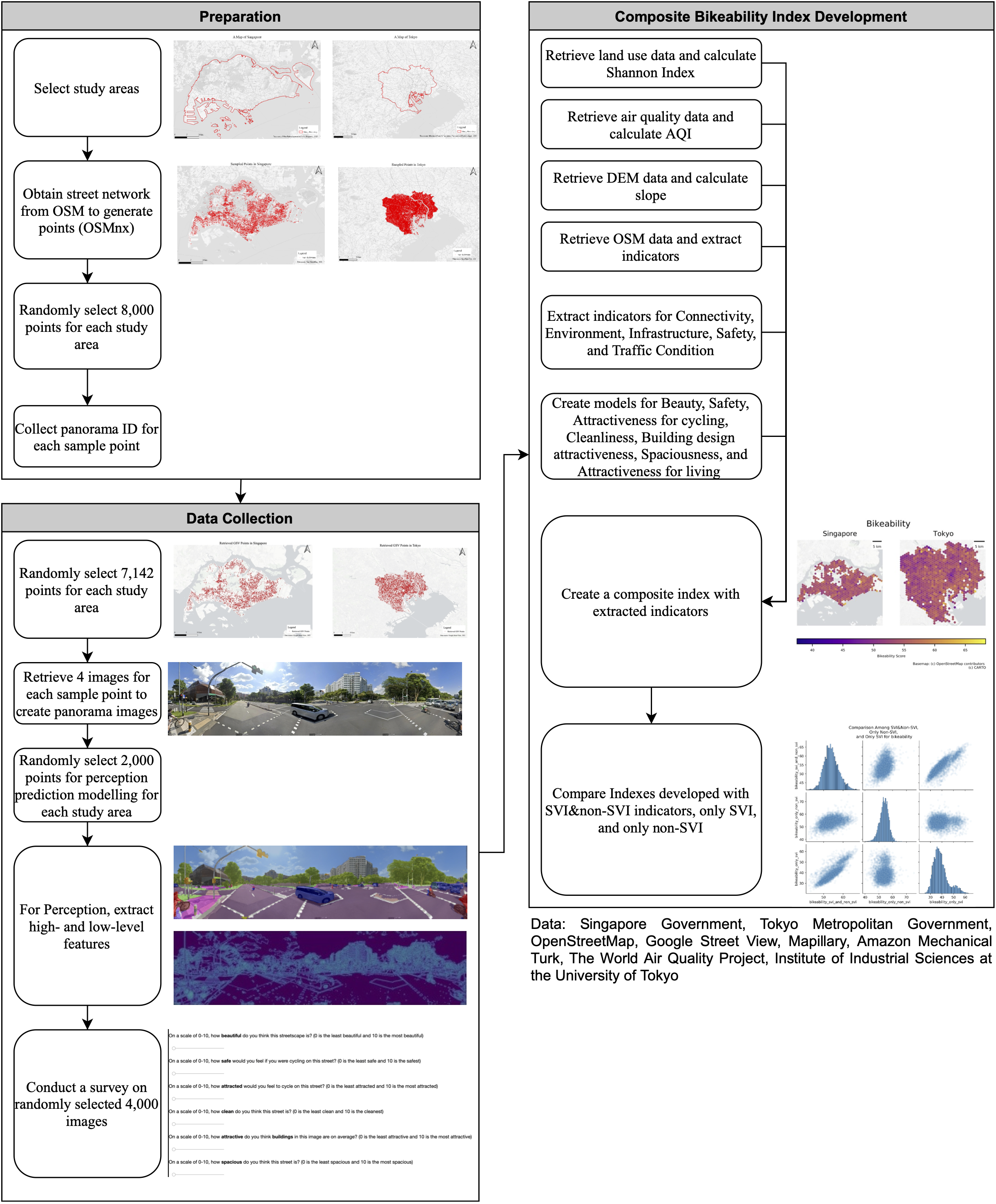}}
  \caption{Illustration of the methodology of this study.}
  \label{methodology:fig:flow_chart}
\end{figure}

Facilitating a critical analysis of the value of SVI over non-SVI counterparts, as one of the principal aims of this study, the following indexes with different types of indicators are designed: 
\begin{enumerate}
  \item Index with SVI indicators and non-SVI indicators
  \item Index with only SVI indicators
  \item Index with only non-SVI indicators
\end{enumerate}
These indexes and their sub-categories were compared with each other to examine how much SVI and non-SVI indicators explain the overall variance. 
Finally, \autoref{methodology:fig:flow_chart} sums up the methodology.

\clearpage

\section{Results}\label{sc:overview}
\subsection{Data collection} %

For OSM data, street network data and point data were retrieved. For street network data, 252,369 street segments and 450,379 street segments were retrieved for Singapore and Tokyo, respectively.
As for point data, 12,640 POIs and 157 mass rail transit stations were collected from Singapore, 75,203 POIs, and 606 stations were collected from Tokyo. 
For SVI, after removing indoor images and grey images from 7,142 images, 5,833 and 6,181 panorama images remained for the two cities.

For LU, each city's open data by the local government was used \citep{tokyo_metropolitan_government_maps_2018, singapore_government_master_2020}. 
Because the categorizations of LU in each city were different, LU categories were harmonized to residential, commercial, and industrial in this study, and other categories were excluded.
AQI data in 2020 are obtained for 5 and 126 stations in the city-state and the Japanese capital, respectively.

\subsection{Extracted indicators and composite index} %
\subsubsection{Connectivity} %
\begin{figure}[!t]
  \centering
  \scalebox{0.1}{\includegraphics[width=6\linewidth]{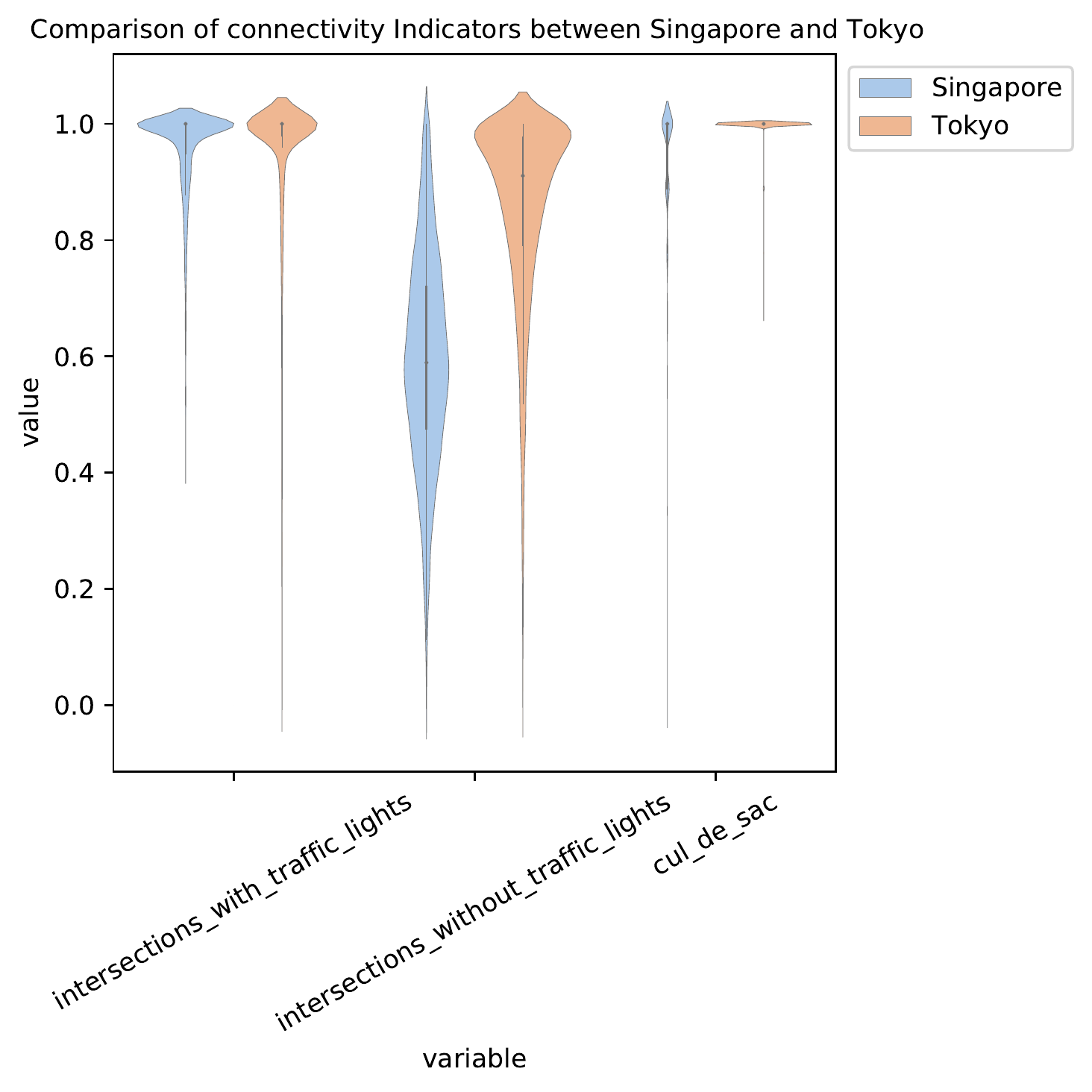}}
  \caption{Distribution of indicators describing connectivity across numerous locations in Singapore and Tokyo.}
  \label{overview:fig:connectivity_by_city}
\end{figure}

Connectivity is assessed based on the number of intersections with lights, intersections without lights, and cul-de-sacs. 
Comparing Singapore and Tokyo, the latter achieves higher scores for connectivity (see
\autoref{overview:tab:summary_statistics} and \autoref{overview:fig:category_by_city}).
Due to a large number of indicators not all of them can be detailed in this paper, but as an example, \autoref{overview:fig:connectivity_by_city} illustrates the distribution of values in this particular indicator.
The score for the number of intersections without traffic lights is much lower for Singapore while the other two indicators have a similar distribution.

\subsubsection{Environment} %
The environment is evaluated based on slope, the number of POIs, land use mix, AQI, and pixel ratios of greenery, buildings, and water, aiming to evaluate the natural and built environment.
Compared to Singapore, Tokyo has a higher mean score for this category (see 
\autoref{overview:tab:summary_statistics} and 
\autoref{overview:fig:category_by_city}).
The distributions of each indicator suggest that Tokyo obtains noticeably higher scores for slope, land use, the pixel ratio of buildings.
Such results reflect Tokyo's relatively flat topography, organic distribution of land uses, and densely built buildings that create more enclosures for cyclists. 
Singapore achieved higher scores in AQI and greenery, which also reflects its strategic environmental management as a garden city \citep{han_singapore_2017,Palliwal.2021}.

\subsubsection{Infrastructure} %
Infrastructure is evaluated based on the type of road and pavement, the width of the road, number of transit facilities, and presence of potholes, street lights, bike lanes, street amenities, utility poles, bike parking spaces, sidewalks, crosswalks, and curb cuts.
This category aims to comprehensively evaluate various elements in the realm of infrastructure.
Compared to Tokyo, Singapore achieves a much higher mean score for this category (see 
\autoref{overview:tab:summary_statistics} and 
\autoref{overview:fig:category_by_city}).
Singapore obtains higher scores in the type of pavement (i.e.\ surface), the presence of street amenity, and, especially, utility poles, which show the nearly opposite result from Tokyo.
This result reflects Tokyo's issue with many utility poles, which not only deteriorates the beauty of the city but also can cause obstruction for pedestrians and cyclists \citep{inajima_koikes_2017}.

\subsubsection{Vehicle-cyclist interaction} %
Vehicle-cyclist interaction was evaluated based on the number of vehicles and speed control devices and the presence of on-street parking and traffic lights/stop signs.
This category aims to assess how safely cyclists interact with vehicles; thus, fewer and slower traffic leads to higher scores.
The mean scores for both cities are also similar, while Singapore has a slightly higher mean and standard deviation  (see 
\autoref{overview:tab:summary_statistics} and 
\autoref{overview:fig:category_by_city}).
The analysis suggests very small variances in the indicators and similar distributions for both cities.
Such a result can potentially be explained: Singapore and Tokyo have similar restrictions on vehicles; for example, on-street parking is strongly discouraged in both cities \citep{barter_parking_2010,russo_environmental_2019}, and the numbers of vehicles per capita are similar: 0.22 and 0.17 \citep{barter_parking_2010,singapore_government_annual_2020}.

\subsubsection{Perception} %
The perception is evaluated based on attractiveness for cycling, spaciousness, cleanliness, building design attractiveness, safety as cyclists, beauty, attractiveness for living.
In this category, we will discuss the survey result, the relationships between perception scores and high- and low-level features, the result of predictive modeling, and the result of inferences from the models.

Surveys on 800 images, 400 images from each city, were conducted on Amazon Mechanical Turk to recruit eight unique participants to rate each image's perception scores mentioned above on a scale of 0 to 10.
Before proceeding further, it is important to assert that while our work strives for a high degree of automation, as in most studies involving machine learning, a portion of the work is manual, i.e.\ the labeling of training data was done through crowdsourcing (Section~\ref{sc:SVI-method}).
However, after the predictive modeling using the data obtained from the survey, the process of determining the scores was automated.

\begin{figure}[!t]
  \centering
  \begin{minipage}[b]{.45\linewidth}
    \centering
    \scalebox{0.1}{\includegraphics[width=10\linewidth]{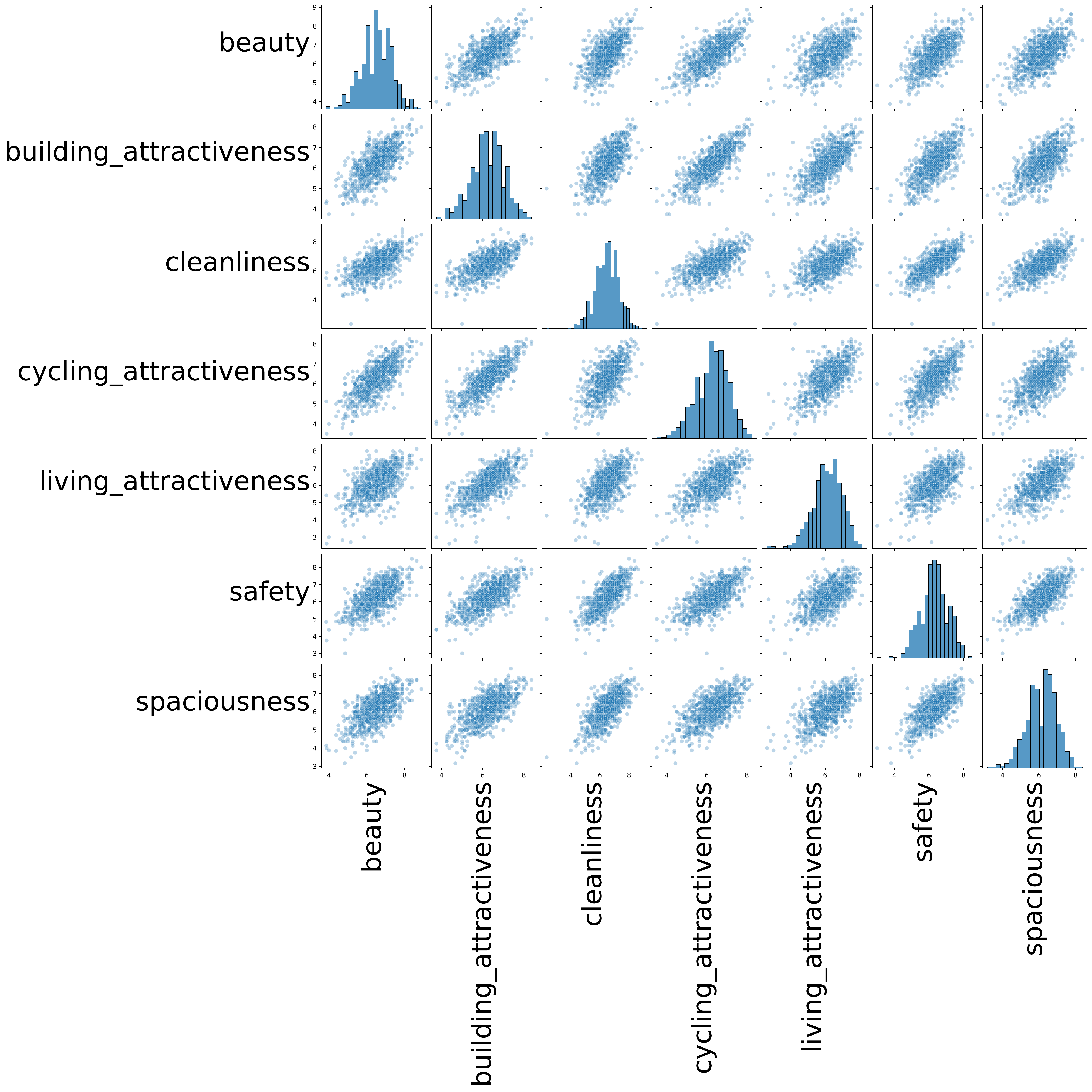}}
    \caption{A scatter plot matrix of the perception scores \\obtained from the survey.}
    \label{overview:fig:survey_scatter_matrix}
  \end{minipage}
  \hspace*{2em}
  \begin{minipage}[b]{.45\linewidth}
    \centering
    \scalebox{0.1}{\includegraphics[width=10\linewidth]{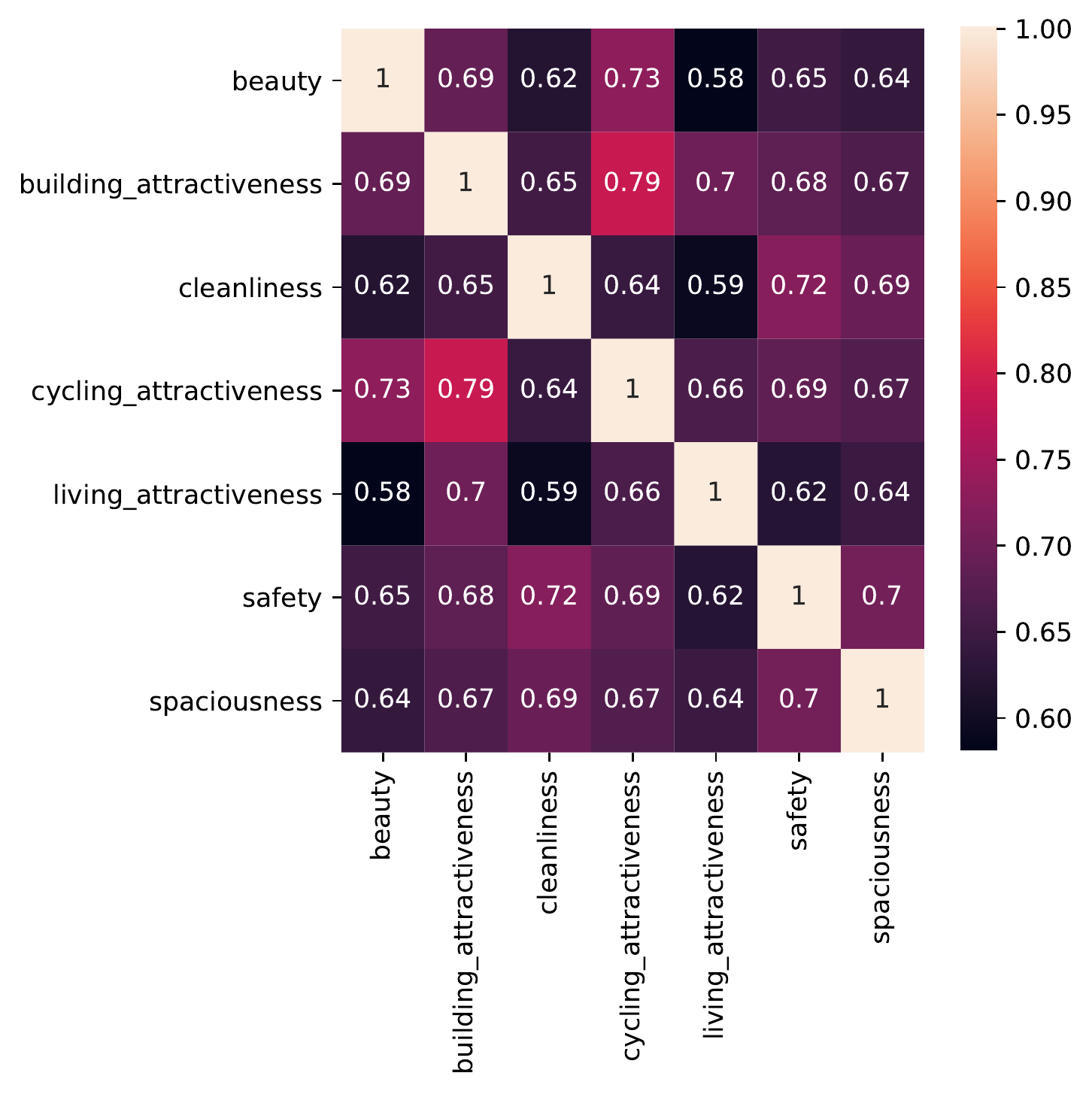}}
    \caption{A correlation matrix of the perception scores \\ obtained from the survey.}
    \label{overview:fig:survey_corr_matrix}
  \end{minipage}
\end{figure}

\begin{figure}[!t]
  \centering
  \scalebox{0.1}{\includegraphics[width=6\linewidth]{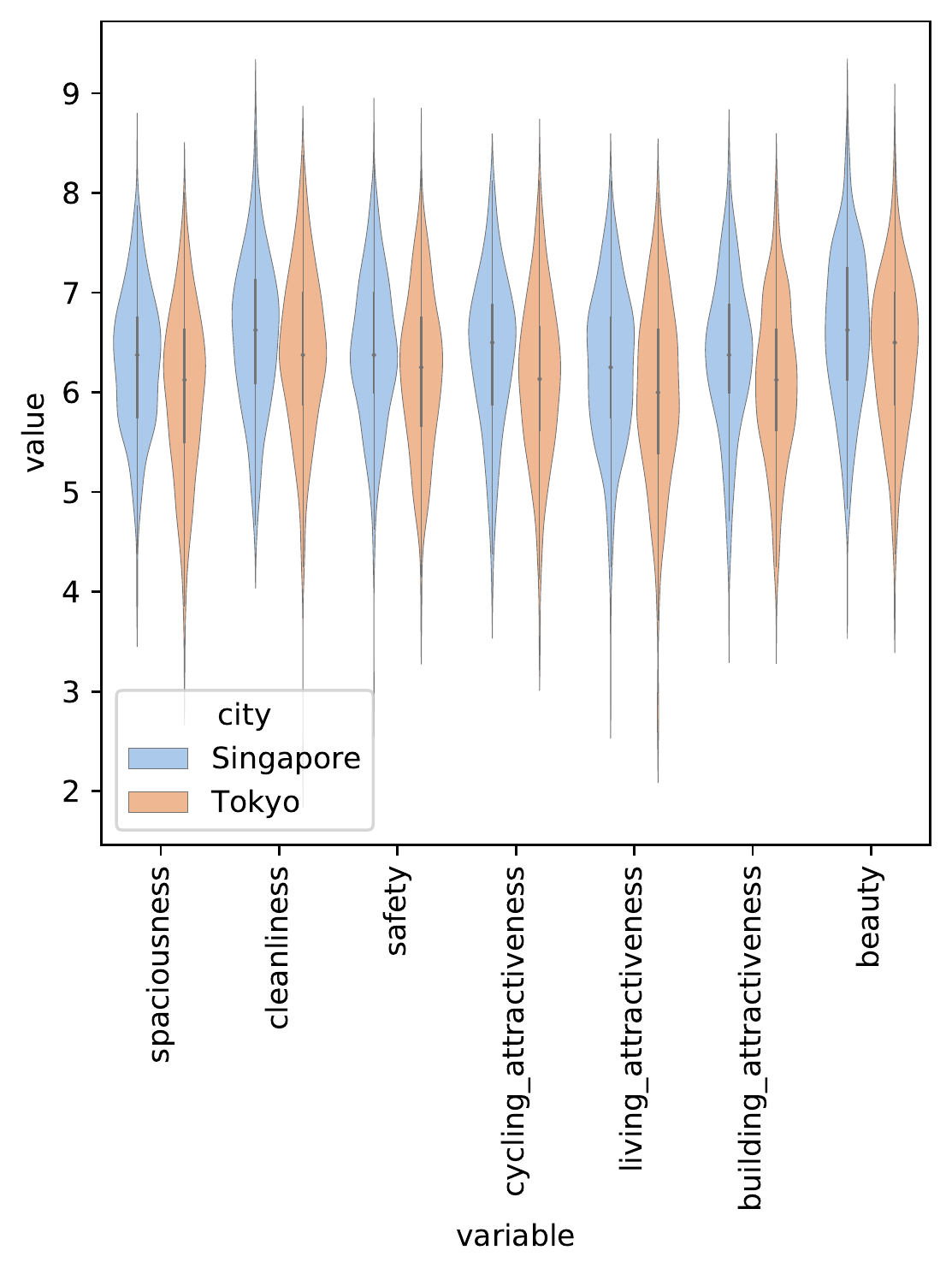}}
  
  \caption{Distribution of seven perception scores obtained from the survey by each city.}
  \label{overview:fig:survey_violion_plot_by_city}
\end{figure}

The results of the survey indicate strong positive correlations among the perception scores, ranging from 0.58 to 0.79 in squared R, and exhibit no visible skewness in the data distribution (see 
\autoref{overview:fig:survey_scatter_matrix} and
\autoref{overview:fig:survey_corr_matrix}).
\autoref{overview:fig:survey_violion_plot_by_city} indicates the results of the survey after excluding outliers, which reveal similar results for all the scores and illustrates that the majority of responses are between three and eight. The figure also suggests that images from Singapore generally had higher scores across all the measures.

Visual features from SVI are extracted through high-level feature extraction (i.e. semantic segmentation, classification, and object detection) and low-level feature extraction (i.e.\ edge detection, blob detection, and HLS statistics).
After labeling each observation below or above the mean value of each feature, Welch's t-test is conducted to find features with statistically significant effects on perception scores based on the labeling.
The total number of extracted features is 519, and the number of scores is seven; therefore, there are 3,633 unique feature-score pairs. 

For beauty, we found that features that contribute to better infrastructures such as curb and terrain contribute to higher scores, while features that damage beauty such as utility poles and junkyards contribute to lower scores.
Also, for low-level features, larger standard deviations of HLS are associated with lower beauty scores.

As for building attractiveness, we realize that a higher pixel ratio of buildings and classification as slums leads to lower scores, implying that participants did not prefer dense buildings. 
On the other hand, classification as residential neighborhood and campus leads to higher scores, suggesting residential and education as potential land uses and typologies that make people feel attracted to buildings. 

The result of cleanliness suggests that classification as campus and higher pixel ratio of curb leads to higher cleanliness scores, which might imply that these places and infrastructures are relatively better maintained. 
On the other hand, classification as junkyard and landfill led to lower cleanliness scores, which is intuitively understandable because of their strong associations with garbage.
As for cycling attractiveness, natural elements such as terrain and vegetation as well as classification as residential neighborhoods leads to higher cycling attractiveness scores, while utility poles and construction sites are associated with lower scores. 
Attractiveness for living obtains a similar pattern to other scores.
A higher pixel ratio of terrain and curb entails higher scores, and classification as slum and landfill causes lower scores.
The result of safety shows that terrain and campus achieve higher scores and that other features such as gas stations, junkyards, and slums were associated with lower safety scores.
Lastly, spaciousness gains higher scores when terrain, curb, and campus are present, while other features, such as utility poles, junkyards, and larger standard deviations of HLS statistics are detrimental.

In this exploratory analysis, some parts of the study by \citet{verma_predicting_2020} are replicated such as low associations among perception scores and edge detection and blob detection.
However, other parts of the study such as the strong influence of cars in object detection could not be reproduced, suggesting that the geography of the study and/or training data may play a role.
Moreover, more features from image classification are found to have an influence on perception scores than other features.

Based on the study by \citet{verma_predicting_2020}, predictive modeling is conducted for each perception score between 0 and 10 with high- and low-level features (see \autoref{methodology:tab:features}).
\autoref{overview:tab:modeling_results} underscores the results of the modeling with different performance metrics, where one can see low mean absolute error (MAE) around 0.65, mean absolute percentage error (MAPE) around 0.1, and root mean squared error (RMSE) around 0.8; however, all the $R^2$ values were below 0.
This means that the models were worse than predicting the constant values regardless of input data.

Although \citet{verma_predicting_2020} achieves $R^2$ as high as 0.66 by using the same visual features, our study could not achieve comparable results.
This discrepancy might be due to the disparate level of data quality in the SVI training set.
After the modeling, we randomly selected SVI images with higher and lower scores and found that some partially grey images and indoor images were still in the set despite the efforts to clean the data prior to the modeling. 
Such noise in SVI data is not often discussed in previous studies using SVI, and this issue remains to be solved in the future~\citep{Biljecki.2021}. 
To further improve the prediction, other visual feature extractions and selection methods need to be explored as well.
\begin{table}[!htp]\centering
\caption{Results of predictive modeling of perception indicators.}\label{overview:tab:modeling_results}
\begin{tabular}{lrrrrr}\toprule
\textbf{Target\_variable} &\textbf{MAE} &\textbf{MAPE} &\textbf{RMSE} &\textbf{R2} \\\midrule
beauty &0.63 &0.10 &0.81 &-0.20 \\
building\_attractiveness &0.63 &0.10 &0.78 &-0.14 \\
cleanliness &0.63 &0.10 &0.79 &-0.12 \\
cycling\_attractiveness &0.65 &0.11 &0.83 &-0.18 \\
living\_attractiveness &0.69 &0.12 &0.87 &-0.10 \\
safety &0.72 &0.12 &0.89 &-0.29 \\
spaciousness &0.67 &0.11 &0.83 &-0.19 \\
\bottomrule
\end{tabular}
\end{table}

\subsubsection{Bikeability} %
\begin{figure}[!t]
  \centering
  \scalebox{0.1}{\includegraphics[width=8\linewidth]{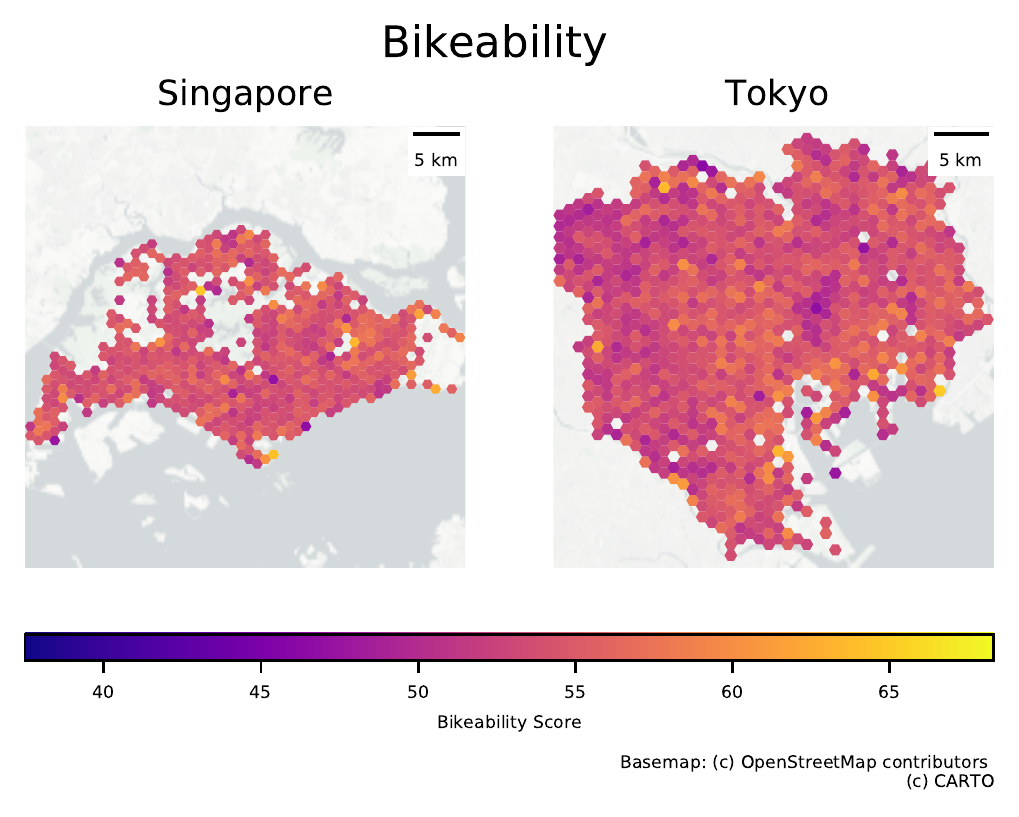}}
  
  \caption{One of the key outputs of this study: maps of bikeability across Singapore and Tokyo, generated from the scores at numerous locations in the two cities.}
  \label{overview:fig:bikeability}
\end{figure}

After incorporating scores across all categories, bikeability scores were calculated at a fine spatial scale (\autoref{overview:fig:bikeability}).
In Singapore, bikeability scores are generally distributed homogeneously barring a few outliers. The results for Tokyo seem to be more heterogeneous, with a lower score in the central area and peripheral areas and higher scores in between.
The distribution of data indicates no skewness in the data, and a comparison between Singapore and Tokyo in \autoref{overview:tab:summary_statistics} hints that Singapore achieved a slightly higher mean value and lower standard deviation than Tokyo. 

\begin{figure}[!t]
  \centering
  \scalebox{0.1}{\includegraphics[width=6\linewidth]{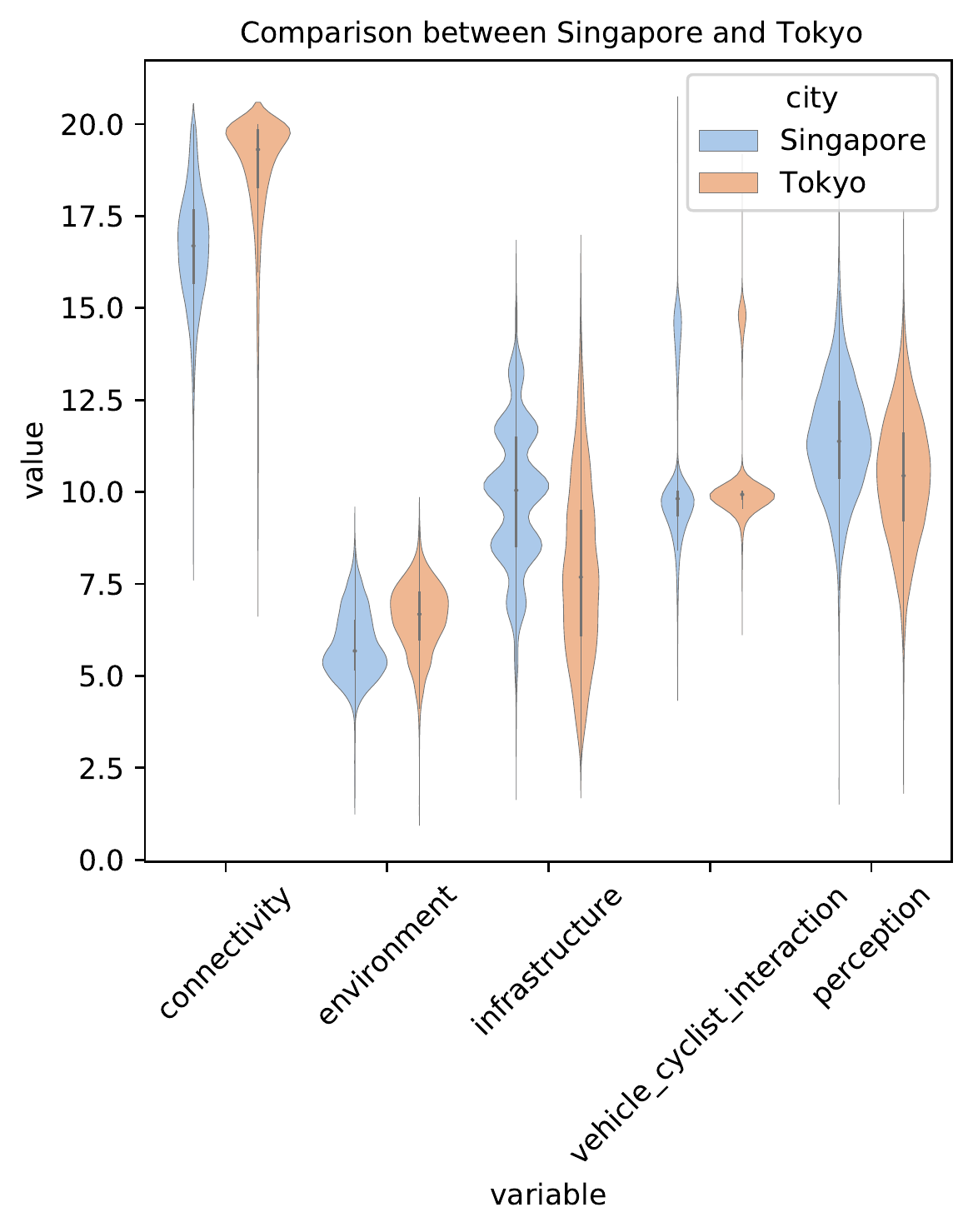}}
  
  \caption{Distribution of scores across categories in Singapore and Tokyo.}
  \label{overview:fig:category_by_city}
\end{figure}

This bikeability index, however, has some issues that need to be exposed. 
Firstly, the perception prediction's result is not entirely reliable. 
Although MAE, MAPE, and RMSE of models turned out to be moderately acceptable, $R^2$ values resulted in negative values, indicating the models are more inaccurate than simply predicting the mean values of the scores.
Low variances among observations in some categories also need to be considered. 
Summary statistics shown in \autoref{overview:tab:summary_statistics} and data distribution shown in \autoref{overview:fig:category_by_city} suggest some categories such as vehicle-cyclist interaction had low standard deviation and skewed data distribution with a few outliers.
This poses a question regarding the appropriate indicators to be used that can differentiate bikeability scores within and between cities because most of the sample points obtain highly similar scores for such categories.
This question needs to be examined by expanding study areas in different contexts to observe whether this phenomenon is simply caused by similar characteristics of streets in the Asian context.

\begin{table}[!htp]\centering
\caption{Summary statistics of the bikeability measures calculated for the study area.}\label{overview:tab:summary_statistics}
\begin{tabular}{llrrr}\toprule
\textbf{Category} &\textbf{City} &\textbf{Mean} &\textbf{Standard deviation} \\\midrule
\multirow{2}{*}{Bikeability} &Singapore &54.26 &3.66 \\
&Tokyo &53.98 &3.73 \\\midrule
\multirow{2}{*}{Connectivity} &Singapore &16.63 &1.59 \\
&Tokyo &18.72 &1.70 \\\midrule
\multirow{2}{*}{Environment} &Singapore &5.83 &0.94 \\
&Tokyo &6.58 &0.94 \\\midrule
\multirow{2}{*}{Infrastructure} &Singapore &9.88 &1.90 \\
&Tokyo &7.86 &2.36 \\\midrule
\multirow{2}{*}{Perception} &Singapore &11.42 &1.57 \\
&Tokyo &10.40 &1.74 \\\midrule
\multirow{2}{*}{Vehicle--cyclist interaction} &Singapore &10.50 &2.12 \\
&Tokyo &10.42 &1.61 \\
\bottomrule
\end{tabular}
\end{table}

\subsection{Comparison between SVI and Non-SVI indicators}
This study compares three types of indexes to answer one of the research questions gauging the value of SVI in bikeability studies.
Besides the comprehensive index developed with both SVI and non-SVI indicators, indexes with only SVI and non-SVI indicators are developed and compared with each other.
Each category's score was recalculated based on the indicator types, and, for bikeability, we compare the indexes by plotting scores developed with different indicators on the same sample points.
Connectivity and perception were excluded from this analysis because these categories only had one type of indicator, only non-SVI indicators, and only SVI indicators, respectively. 

\begin{figure}[!t]
  \centering
  \begin{minipage}[b]{.45\linewidth}
    \centering
    \scalebox{0.1}{\includegraphics[width=9\linewidth]{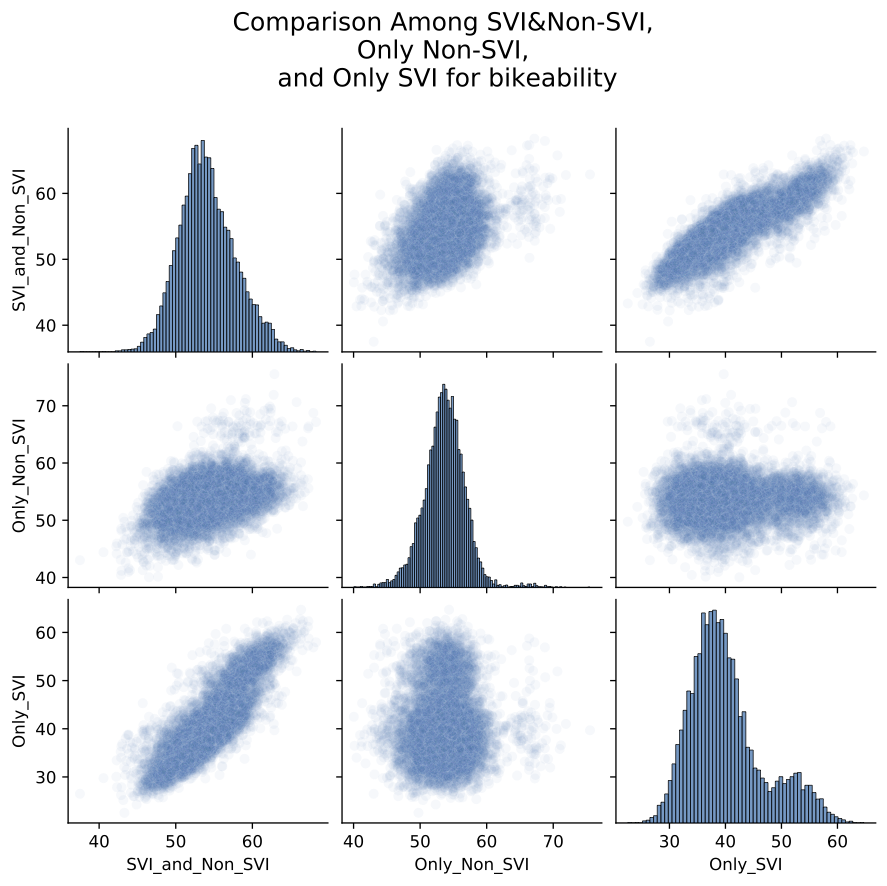}}
    \caption{A scatter plot matrix of bikeability score \\ with only SVI and non-SVI indicators.}
    \label{overview:fig:bikeability_scatter_plot}
  \end{minipage}
  \hspace*{2em}
  \begin{minipage}[b]{.45\linewidth}
    \centering
    \scalebox{0.1}{\includegraphics[width=9\linewidth]{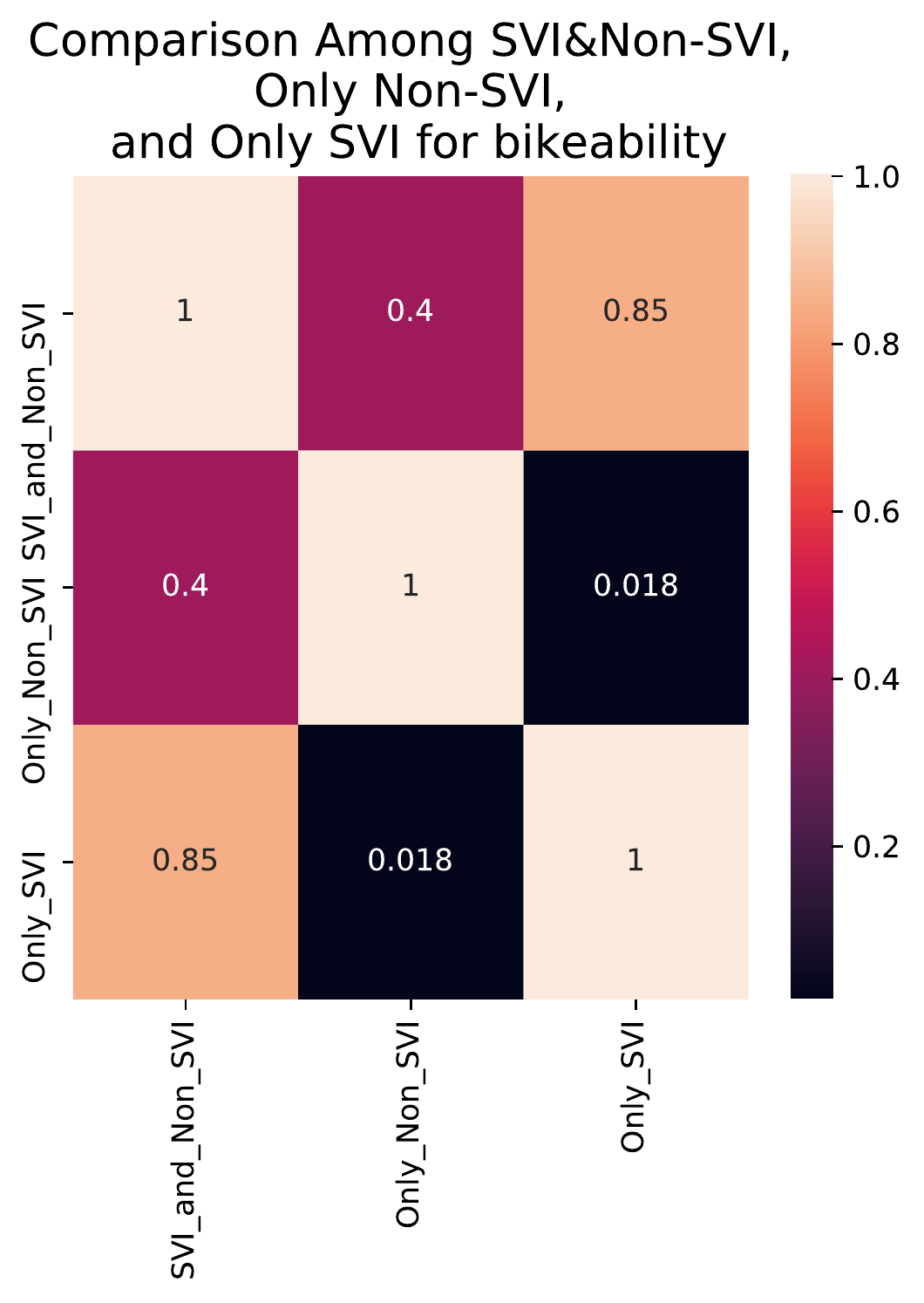}}
    \caption{A correlation matrix of bikeability score \\ with only SVI and non-SVI indicators.}
    \label{overview:fig:bikeability_corr_matrix}
  \end{minipage}
\end{figure}

Indexes of the environment developed with SVI and non-SVI indicators are not correlated with each other and that non-SVI indicators are more strongly correlated with the index developed with both of the indicators. 
On the other hand, SVI indicators had more influence on infrastructure and vehicle-cyclist interaction. 
After combining all the categories, bikeability scores were compared among different indicator types, and SVI indicators turned out to have a stronger correlation with the index with both indicator types and lower kurtosis than non-SVI indicators (see \autoref{overview:fig:bikeability_corr_matrix} and \autoref{overview:fig:bikeability_scatter_plot}). 
Although this result is not a surprise given the larger number of SVI indicators than non-SVI indicators, this result indicates the potential of SVI indicators to be used alone to evaluate bikeability because of its high correlation, $R^2$ of 0.85, with the index with both indicators and capability to explain most of the variance.

Although it was found possible to estimate bikeability only with SVI indicators, some relevant non-SVI indicators are either difficult or impossible to replace with SVI.
For example, transit facilities, POIs, and land uses are not impossible but complicated to obtain from SVI and are much more straightforward to collect from OSM and other data sources, which are anyway frequently available.
Also, the slope and air quality are unnecessarily challenging to estimate from SVI, and other data sources are much less involved to collect and much more accurate than it would be if estimating them from SVI. 
Therefore, with these outcomes in mind, it is more beneficial for urban planners and researchers to combine both SVI and non-SVI indicators to assess bikeability, complementing the best of the two worlds.

\section{Challenges and future directions}
\subsection{Data quality}
This study faced several challenges.
One of such challenges was the quality of data.
This issue surfaced in data sources, such as SVI, OSM, and AQI.
GSV was used to collect SVI in this study, and the sample points of collected GSV panorama images were randomly selected to cover the entire study areas with a limited number of sample points of GSV that can be obtained for free.  
A study by \citet{kim_decoding_2021} reported that larger sampling intervals lead to larger variances of elements that can be extracted from SVI.
Therefore, one of the limitations of this study is possible biases and larger variances of data introduced by random sampling intervals when conducting random sampling. 
Future studies can define shorter sampling intervals to reduce the possible bias. 
Another possible bias is the perspective of SVI.
Because SVI is collected from vehicles, it might not always represent the typical view of bicyclists, especially for wide roads. 
This limitation is also faced by other previous studies on walkability that as well use 360-degree panoramas for assessment, as they aim to gauge what pedestrians perceive \citep{wakamiya_pleasant_2019,yencha_valuing_2019,nagata_objective_2020,villeneuve_comparing_2018,li_investigating_2018,zhang_walking_2018,weld_deep_2019}.
In this study, the result showed that cycling paths are not highly prevalent in both Singapore and Tokyo; thus, the panorama views possibly present most of the actual cycling perspectives. 
Although SVI on cycling paths is not widely available currently, future studies might be able to work on SVI on cycling paths with the rise of crowdsourced SVI services, such as Mapillary.

As for OSM data, while in general of high quality for both locations, its data completeness became a bottleneck when evaluating some particular indicators (e.g.\ traffic speed and the number of traffic lanes) and resulted in their eliminations, so future studies should explore different data sources to collect them.

AQI data are available in both cities, but the number of AQI stations in Singapore, five, was far fewer than Tokyo, 126, and such low spatial granularity of data might have affected the result.
Therefore, future studies can incorporate predictive modeling to estimate AQI for each sample point by using traffic data, meteorological data, and land use data \citep{chen_land_2010}.

Low variances among sample points for some indicators were found as well.
This is potentially problematic because indicators with extremely low variances cannot differentiate sample points; thus, this needs to be examined more by investigating more diverse sets of cities.

Perception modeling was another challenge faced in this study due to the nature of the approach.
The comprehensive survey was conducted on a service on which the majority of participants are not residents of the study areas, which may be advantageous but also detrimental.
On the one hand, survey results might have been different if they were conducted with residents in the study areas \citep{difallah_demographics_2018}.
To mitigate the potential bias, future studies can utilize different survey services that can specify the residence of the participants.
On the other hand, having participants of a cross-city study residing in neither study area may mitigate the bias of residents living in one city but not being familiar with the other one.

\subsection{Required skills}
Because the data collection and processing was conducted using Python, this method requires users to have a moderately advanced understanding of programming.
Computational power is also a challenge because this study's method uses CV techniques that require graphics processing capabilities, which are not available widely.
To mitigate these limitations, future studies may consider developing a GUI software and an API to allow users to input data and assess bikeability.

\subsection{Development of the index}
This study selected the indicators based on the systematic review.
However, purposes of cycling and socio-economic characteristics might affect the indicator selection and weight assignment, as examined by \citet{arellana_developing_2020}.
Moreover, a bikeability assessment based on route simulations between origin and destinations called capability-wise walkability score has been proposed by a previous study \citep{blecic_capability-wise_2021}.
Thus, the incorporation of such new designs of bikeability index might be able to enhance our method as well.

\section{Conclusion}\label{sc:conclusion}
In previous studies, it has been a challenge to make bikeability assessment scalable while keeping a good balance of objectivity and subjectivity, and minimizing the work required. 
The few studies that have used SVI and CV techniques to automate the process and increase the geographic coverage rather assessed limited aspects of bikeability and have not done so critically, nor have they investigated the inclusion of multiple cities.

We advanced the comprehensive assessment of bikeability using street view imagery and computer vision.
The contributions of our study are the creation of an exhaustive bikeability index inspired by previous studies with SVI indicators extracted with CV techniques as the major data source, novel exploration of automatable subjective assessment of bikeability, comparison of SVI and non-SVI indicators for the first time, and a novel investigation of the potential of using SVI indicators independently for bikeability assessment. 

Non-perception SVI indicators were extracted using semantic segmentation and object detection, and perception SVI indicators were predicted by training models with survey results as target variables and visual features extracted from SVI as independent variables.
Bikeability indexes that range from 0 to 100 were developed in Singapore and Tokyo and compared with each other, which resulted in slightly higher bikeability scores in Singapore, 54.26 on average, than in Tokyo, 53.98 on average.
A thorough comparison between SVI and non-SVI indicators was made to examine which has more influence when predicting conditions and the appeal of cycling.
SVI indicators turned out to have a much stronger correlation with the estimated bikeability index, an $R^2$ of 0.85, outperforming non-SVI indicators, which had an $R^2$ of 0.4.
However, the usefulness of the latter should not be discounted.

In summary, the takeaways of this research are:
\begin{itemize}
    \item This study has demonstrated that we can use CV techniques and SVI to comprehensively assess bikeability within and among cities. The paper details the design and implementation of a bikeability index that relies on SVI and CV, which is calculated at both a fine spatial scale and aggregated at the level of a city. The study asserts that this index at least supplements traditional instruments used in this research domain.
    \item Indicators that are derived from SVI dwarf those that are computed from non-SVI counterparts. A large portion  of the variance in the overall index was explained by SVI indicators, overshadowing those prominent in orthodox mechanisms used hitherto.
  \item SVI may potentially be used on its own to assess bikeability in the built environment. However, as elaborated in the previous sections, this study faced several challenges that, despite the convincing usability of SVI, may not make taking this independent route always viable. Thus, future studies need to ameliorate several practical issues. Further, the comparative advantage comes at a price --- it is more difficult to obtain SVI indicators in comparison to the non-SVI counterparts. Therefore, despite the relative usability and independence of either, it might be beneficial to use both to assess bikeability, complementing their pros and cons.
  \end{itemize}

Based on the findings from this study, future research should focus not only on overcoming the challenges discussed in the previous section but also on further enhancing the index, which advances the state of the art but may nevertheless benefit from further work. 
The index can be improved by expanding the scope of SVI indicators to be included and also improving the indicator weights according to cyclists' preferences.
The enlargement of the indicators can be done by building CV models to extract more indicators, such as road type, land use, and type of pavement. 
Some studies have explored creating weights based on cyclists' preferences through surveys, creating different weights for different types of cyclists \citep{arellana_developing_2020}.
Such improvement of the weights can benefit better decision-making in urban planning according to the demographics in target areas, thus, future studies should also incorporate such methods.
Another direction for future work would be coupling the developed index with instances introduced to assess other urban aspects such as walkability and livability \citep{Zhao.2020gpb,Benita.2020}, to investigate relationships or complement them.

\section*{List of acronyms}
\begin{tabular}{ll}%
AQI & Air Quality Index \\
CV & Computer Vision \\
DEM & Digital Elevation Model \\ 
HSL & Hue-Saturation-Lightness \\
IC & Image Classification \\ 
LU & Land Use \\ 
MAE & Mean Absolute Error \\ 
MAPE & Mean Absolute Percentage Error \\
GSV & Google Street View \\
OD & Object Detection \\ 
OSM & OpenStreetMap \\
POI & Point of Interest \\
RMSE & Root Mean Squared Error \\
SS & Semantic Segmentation \\
SVI & Street View Imagery \\
V--C Interaction & Vehicle--Cyclist Interaction \\
\end{tabular}

\section*{Acknowledgments}
We appreciate the comments by Jeffrey Ho and the design contribution by April Zhu (National University of Singapore).
We thank the members of the NUS Urban Analytics Lab for the discussions and the reviewers for their suggestions.
The data sources used in this study are gratefully acknowledged.
This research is part of the project Large-scale 3D Geospatial Data for Urban Analytics, which is supported by the National University of Singapore under the Start Up Grant R-295-000-171-133.
This study has received an exemption from the Institutional Review Board (IRB) of the National University of Singapore under the reference code NUS-IRB-2021-29.

\bibliographystyle{cas-model2-names}

\bibliography{references}

\end{document}